\documentclass[runningheads]{llncs}

\usepackage{eccv}

\usepackage{eccvabbrv}

\usepackage{graphicx}
\usepackage{booktabs}

\usepackage[accsupp]{axessibility}  

\usepackage{hyperref}

\usepackage{orcidlink}

\usepackage[utf8]{inputenc} 
\usepackage[T1]{fontenc}    
\usepackage{url}            
\usepackage{amsfonts}       
\usepackage{nicefrac}       
\usepackage{microtype}      
\usepackage{xcolor}         
\usepackage{subcaption}
\usepackage{caption}
\usepackage{float}
\usepackage{amsmath}
\usepackage{amssymb}
\usepackage{array} 
\usepackage{adjustbox}

\providecommand{\citep}[1]{\cite{#1}}
\providecommand{\citet}[1]{\cite{#1}}
\newcommand\blfootnote[1]{%
  \begingroup
  \renewcommand\thefootnote{}\footnote{#1}%
  \addtocounter{footnote}{-1}%
  \endgroup
}

\usepackage{amsmath,amsfonts,bm}

\def\eqref#1{equation~\ref{#1}}

\def\1{\bm{1}}

\DeclareMathAlphabet{\mathsfit}{\encodingdefault}{\sfdefault}{m}{sl}
\SetMathAlphabet{\mathsfit}{bold}{\encodingdefault}{\sfdefault}{bx}{n}

\newcommand{\red}[1]{{\color{red}#1}}

\newcommand{\ra}[1]{\renewcommand{\arraystretch}{#1}}

\begin{document}

\title{Lifelong Learning of Video Diffusion Models\texorpdfstring{\\}{ }From a Single Video Stream}


\author{
    Jason Yoo\inst{1} \and
    Yingchen He\inst{1} \and
    Saeid Naderiparizi\inst{1} \and
    Dylan Green\inst{1} \and \\
    Gido van de Ven\inst{2} \and
    Geoff Pleiss\inst{1} \and
    Frank Wood\inst{1}
}

\authorrunning{J.~Yoo et al.}

\institute{
    Univeristy of British Columbia, Vancouver, Canada \and 
    University of Groningen, Groningen, Netherland
\\
\email{jasonyoo0116@gmail.com}\\
}

\maketitle
\setcounter{footnote}{0}

\begin{abstract}
Video diffusion models can enable embodied agents to anticipate plausible futures from the recent past, but they are typically trained offline on curated datasets—a mismatch with the agents' learning setup at deployment: online, from a single video stream that sequentially outputs one frame at a time.
We bridge this training gap and demonstrate that training autoregressive video diffusion models from such a stream, resembling the experience of embodied agents, is not only possible but can also perform comparably to standard offline training given the same number of gradient steps.
We find that this robustness to video stream autocorrelation and nonstationarity can be achieved using experience replay methods that retain a subset of the video stream.
To support training and evaluation in this setting, we introduce five new datasets for streaming lifelong generative video modeling: \textit{Lifelong Bouncing Balls (O)}, \textit{Lifelong Bouncing Balls (C)}, \textit{Lifelong 3D Maze}, \textit{Lifelong Drive}, and \textit{Lifelong PLAICraft}, each consisting of one million consecutive frames from environments of increasing complexity.
Together, our datasets and experiments lay the groundwork for video generative models and world models that continuously learn from single-sensor video streams rather than fixed datasets.
\end{abstract}

\section{Introduction}
\label{sec:intro}

There are a plethora of names -- lifelong learning, continual learning, streaming inference -- given to a central desideratum of artificial intelligence (AI) systems: the ability to learn from a single continuous autocorrelated stream of data.  However named, our community has long sought models and algorithms that learn in a fundamentally human way; from birth to death, learning as we live.

Modern AI systems do not learn in this manner but instead rely on an effective compromise: stochastic gradient descent from data streams made up of independently and identically distributed (i.i.d.)\ samples.
\blfootnote{Datasets and code are available at \url{https://github.com/plai-group/lifelong-vdm}.}
Language models \citep{touvron2023llama,mukherjee2023orca}, world models \citep{hafner2020dream, hafner2021mastering,videoworldsimulators2024, agarwal2025cosmos}, and video models \citep{harvey2022flexible,ho2022video, bar2024lumiere} are trained on random batches of short temporally correlated segments,
an approach that preserves short-range autocorrelations while approximating i.i.d.-ness through permutations.
%
While effective, this approximately i.i.d.\ learning paradigm does not offer a satisfying mechanism for updating models when new data arrive.
Existing model updating techniques that are often used in practice---%
training from a checkpoint
on the union of old and new data \citep{ash2020warm},
fine-tuning on just the new data \citep{xu2023parameter},
or training completely anew from scratch \citep{ren2021survey} --
do not operate in the training regime from which humans learn and suffer from problems such as high computation cost and forgetting \citep{verwimp2024continual}.

\begin{figure*}[t!]
    \centering
    \newcommand{\rowlabel}[1]{\fontsize{6pt}{6pt}\selectfont {#1}}
    \begin{tabular}{@{}m{0.875\linewidth}@{\hspace{0.4em}}m{0.1\linewidth}@{}}

    \begin{minipage}[t]{\linewidth}
        \vspace{0pt}
        \centering
        \begin{subfigure}[t]{.495\linewidth}
            \centering
            \includegraphics[width=\linewidth, height=4.25cm, keepaspectratio=false]{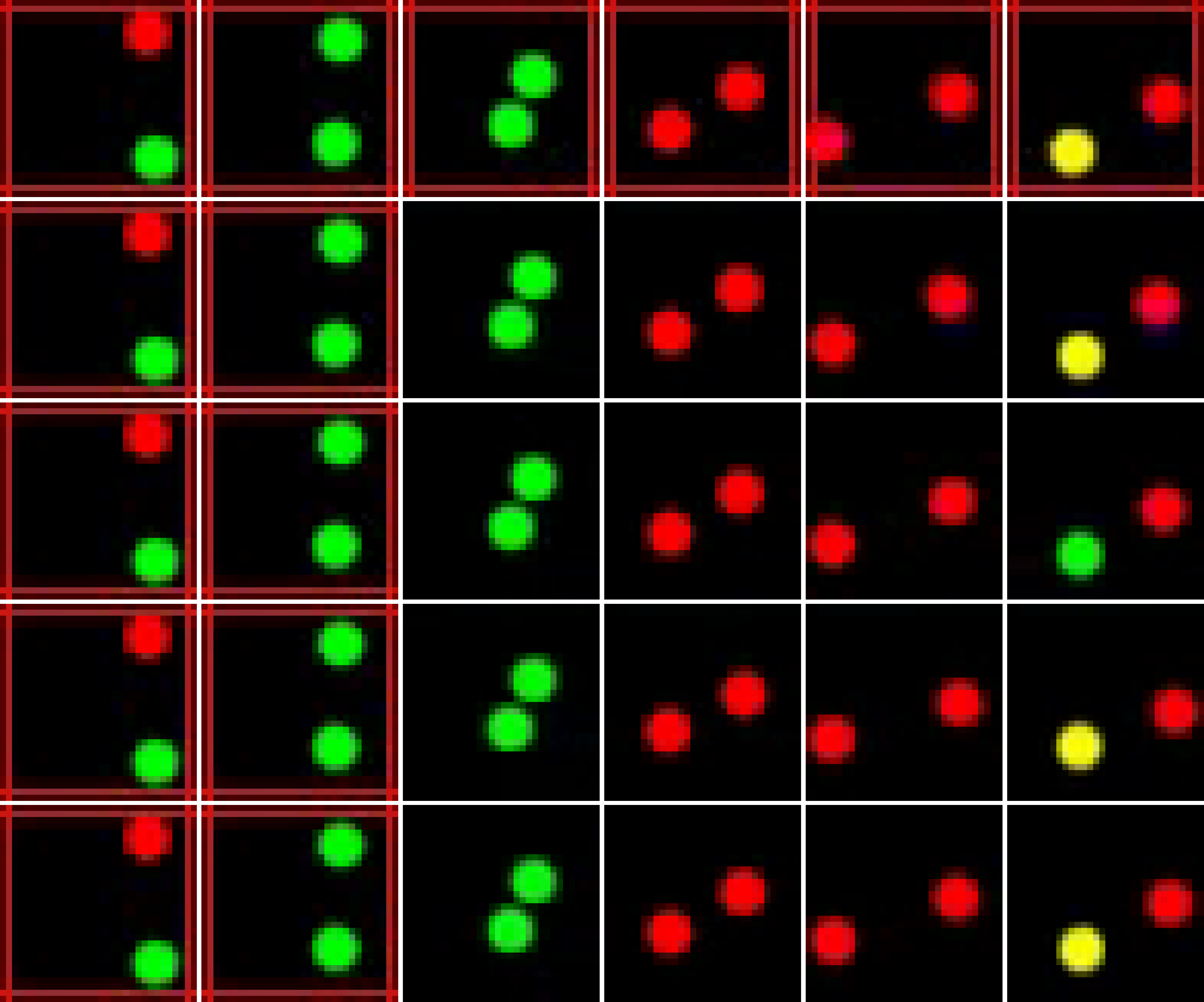}
            \caption{\textit{Lifelong Bouncing Balls} videos.}
            \label{fig:ball_examples}
        \end{subfigure}
        \hfill
        \begin{subfigure}[t]{.495\linewidth}
            \centering
            \includegraphics[width=\linewidth, height=4.25cm, keepaspectratio=false]{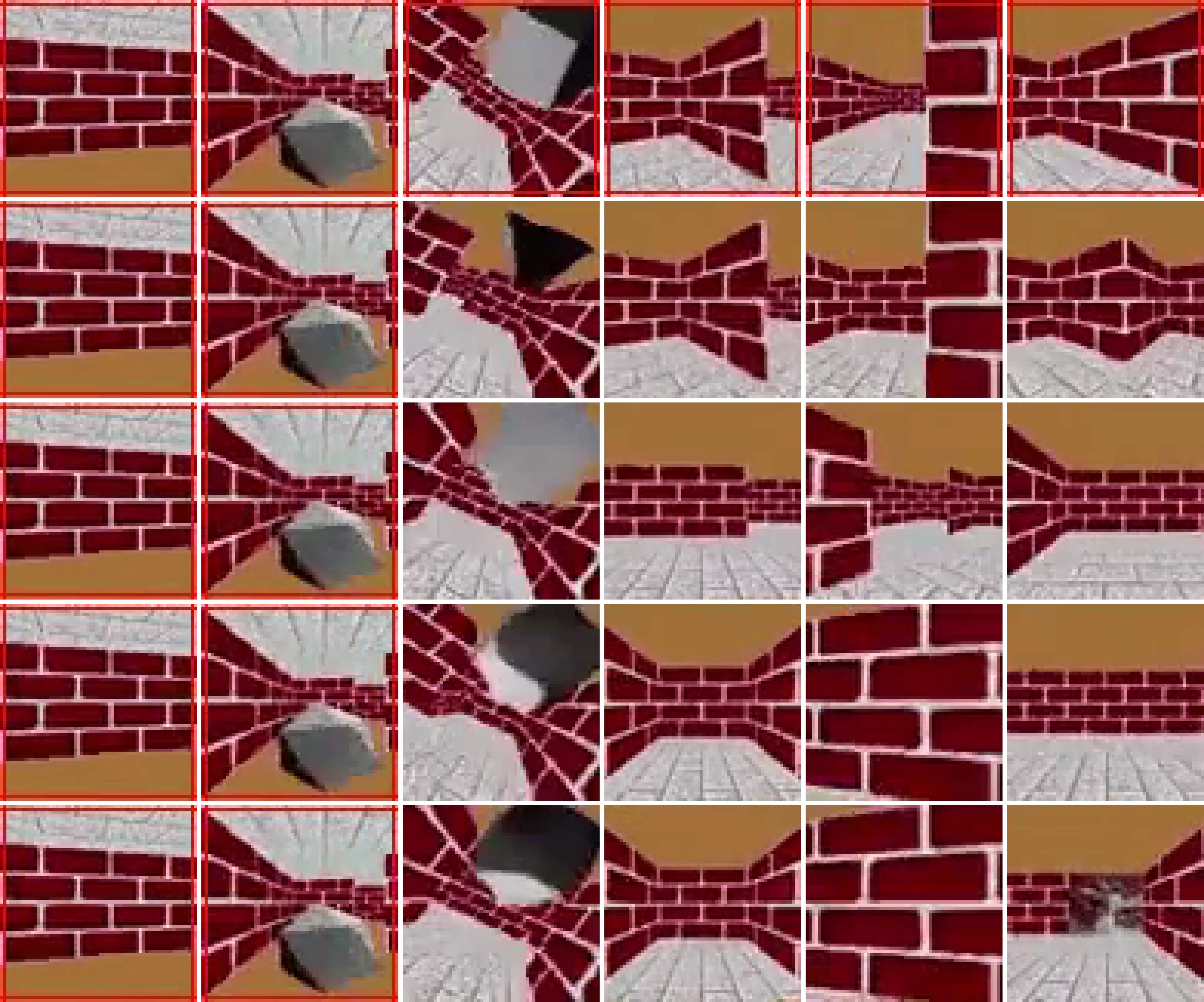}
            \caption{\textit{Lifelong 3D Maze} videos.}
            \label{fig:maze_examples}
        \end{subfigure}
    \end{minipage}
    &
    \begin{minipage}[t]{\linewidth}
    \vspace*{0.75em}
    \rowlabel{Ground Truth} \\ [1.25em]
    \rowlabel{Offline Learning Sample 1} \\ [0.8em]
    \rowlabel{Offline Learning Sample 2} \\ [0.8em]
    \rowlabel{Lifelong Learning Sample 1} \\ [0.8em]
    \rowlabel{Lifelong Learning Sample 2}
    \end{minipage}
    \\
    \begin{minipage}[t]{\linewidth}
        \vspace{0pt}
        \centering
        \begin{subfigure}[t]{.495\linewidth}
            \centering
            \includegraphics[width=\linewidth, height=4.25cm, keepaspectratio=false]{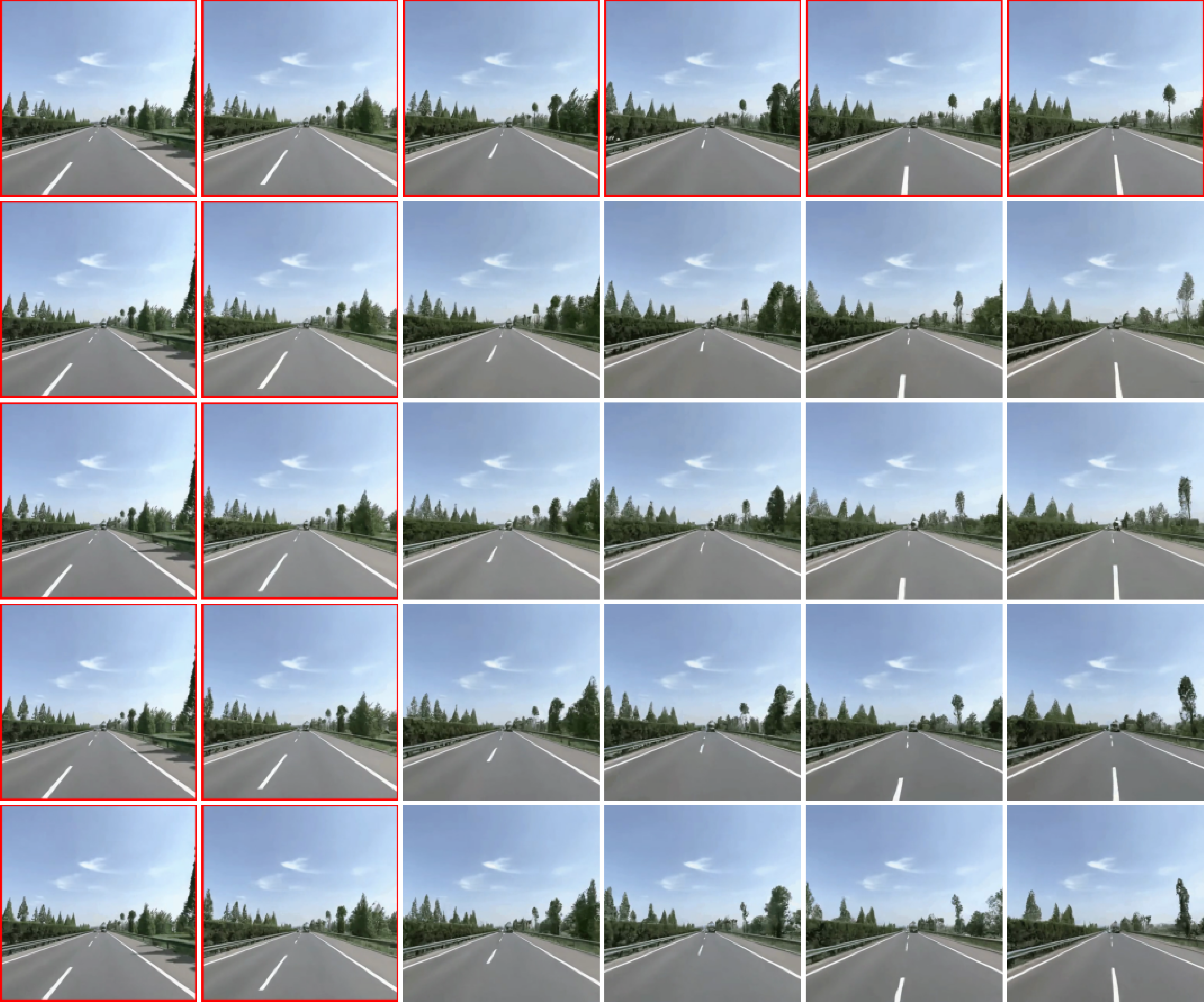}
            \caption{\textit{Lifelong Drive} videos.}
            \label{fig:drive_examples}
        \end{subfigure}
        \hfill
        \begin{subfigure}[t]{.495\linewidth}
            \centering
            \includegraphics[width=\linewidth, height=4.25cm, keepaspectratio=false]{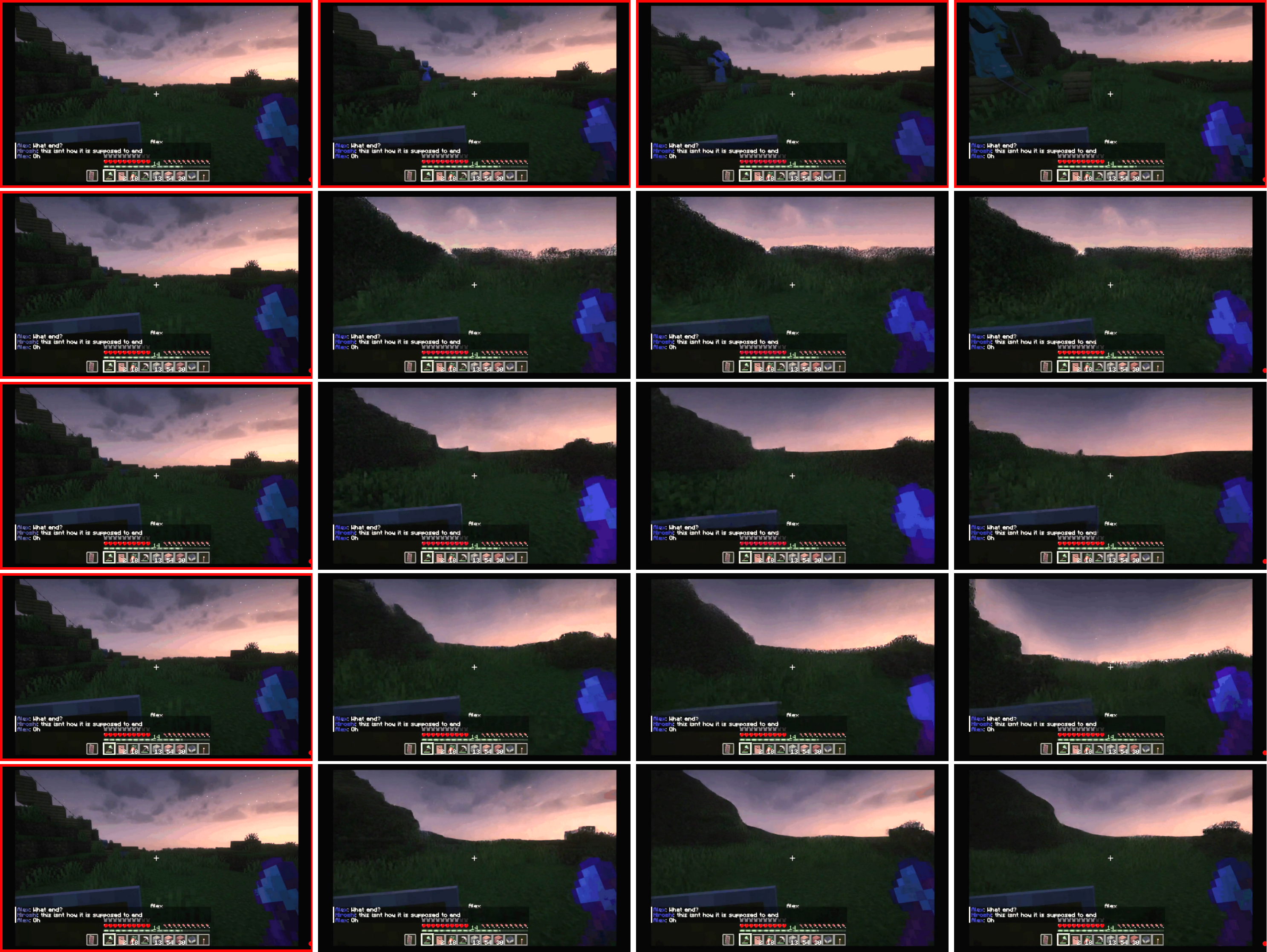}
            \caption{\textit{Lifelong PLAICraft} videos.}
            \label{fig:minecraft_examples}
        \end{subfigure}
    \end{minipage}
    &
    \begin{minipage}[t]{\linewidth}
    \vspace*{0.75em}
    \rowlabel{Ground Truth} \\ [1.25em]
    \rowlabel{Offline Learning Sample 1} \\ [0.8em]
    \rowlabel{Offline Learning Sample 2} \\ [0.8em]
    \rowlabel{Lifelong Learning Sample 1} \\ [0.8em]
    \rowlabel{Lifelong Learning Sample 2}
    \end{minipage}
    \\
    \end{tabular}
    \caption{Ground truth video frames (1st row of each subfigure), offline learned models' generated videos (2nd and 3rd rows), and lifelong learned models' generated video frames (4th and 5th rows) for our datasets.
    The leftmost columns highlighted in \red{red} show the frame conditioned upon the model.
    Videos generated by lifelong learned models trained with experience replay are diverse, visually plausible, and comparable to offline models in visual fidelity.
    Figures are best viewed zoomed in.
    }
    \label{fig:examples}
\end{figure*}

Alternatively, SGD on autocorrelated data streams is considered by some to be a viable candidate for human-like lifelong learning \citep{Lillicrap2020}.
While there is a raft of work indicating that gradient-based learning on autocorrelated data is possible \citep{duchi2012ergodic,johansson2010randomized,ram2009incremental},
folk wisdom holds that this setup is difficult, impractical, and failure-prone, especially for deep networks.
The optimization community has developed numerous methods for alleviating the effects of data stream temporal dependencies \citep{kowshik2021streaming,godichon-baggioni2023learning,chang2022distributed}, suggesting the existence of problems with learning from autocorrelated data streams.
The Bayesian learning community has proposed numerous continual learning approaches through posterior updating
\citep{bartlett2011deplump,broderick2013streaming,naesseth2019elements,pmlr-v161-beronov21a}, though they lack practical scaled results.
These bodies of work suggest a strong desire to avoid non-i.i.d. learning,
making it reasonable to assume that lifelong learning of video models from a single autocorrelated video stream might be highly challenging.\footnote{This paper uses the term \textit{lifelong learning} interchangeably with \textit{streaming learning} or \textit{continual learning}, where a learner receives a single data item (ex. video frame) per timestep. This is consistent with how living entities perceive the world.}

The main contribution of this paper is an empirical proof-of-concept showing that video diffusion models can learn in a lifelong manner from a single autocorrelated video stream while obtaining similar performance to standard (offline) i.i.d. training.
We demonstrate successful lifelong learning even on highly nonstationary partially-observable 3D domains across multiple datasets, architectures, parameter sizes, and optimizers.
Contrary to what one might expect, addressing data stream autocorrelation and nonstationarity does not necessitate a complex setup:
Experience replay effectively alleviates catastrophic forgetting seen when training on only the latest video frames, though it is less critical in the absence of significant data stream nonstationarity.
We find that this well-established technique can make lifelong- and offline-trained models qualitatively indistinguishable given the same number of gradient steps and batch size.

As an additional contribution, we introduce five single video lifelong learning datasets with varying levels of temporal correlation, repetitiveness, perceptual complexity, and non-stationarity.
Even with an academic computational budget,
we observe stable learning and reliable short-range extrapolations on every dataset.
As video modeling is a component of world modeling,
our findings have the potential to open up new life-like streaming approaches to learning, planning, and control in embodied agents.

\section{Background}
\label{sec:background}

\paragraph{Video Diffusion Models.}
Video Diffusion Models (VDMs) are a class of generative models capable of synthesizing high-quality, temporally consistent videos \citep{voleti2022mcvd,ho2022video,harvey2022flexible,hoppe2022diffusion,green2024semantically,blattmann2023align,blattmann2023stable,lu2024vdt,videoworldsimulators2024}.
Rooted in the principles of denoising diffusion, video diffusion models progressively refine random noise into coherent video frames across a series of iterative denoising steps.
Extending image diffusion models to videos requires capturing temporal dependencies between frames while preserving single-frame quality, which existing work typically achieves using attention mechanisms.

\paragraph{Lifelong Learning.}

The goal of lifelong or continual learning is enabling models to continuously learn from new data with minimal forgetting of what was learned before \citep{delange2022continual,wang2024comprehensive,van2024continual,yoo2022bayespcn}. Approaches to lifelong learning include using regularization to penalize changes to parts of the network that encode previously learned information \citep{kirkpatrick2017overcoming,zenke2017continual,li2017learning}, improving the plasticity or stability of the optimization algorithm \citep{dohare2024loss,hess2023two,yoo2024layerwise}, and enforcing the encoding of different tasks in orthogonal parts of the model \citep{rusu2016progressive,serra2018overcoming,zeng2019continual}. Another popular approach is replay, whereby the model revisits samples representative of past data along with the current training data \citep{robins1995catastrophic}. Such replayed samples can be obtained from generative models \citep{shin2017continual}, but often they are sampled from a memory buffer containing past training data, an approach referred to as experience replay \citep{chaudhry2019tiny,rolnick2019experience,buzzega2020dark,arani2022learning}.

Most work on neural network lifelong learning has focused on the highly simplified problem setting where a classification model is trained on a sequence of non-overlapping tasks, with each task seen only once. Often the model can train on each new task until convergence (referred to as the offline setting), although some works only allow a single pass over the data of each task (the online setting) \citep{aljundi2019gradient,chen2020mitigating}. Beyond classification, lifelong learning research has explored the continual training of generative models, including GANs, VAEs and diffusion models for image generation \citep{zhai2019lifelong,lesort2019generative,egorov2021boovae,smith2024continual}. However, no work thus far has explored lifelong learning of video diffusion models. The lifelong learning community has also shown interest in moving beyond learning in a strictly task-based manner. To create data streams with more complex temporal correlations, blurry task boundaries \citep{bang2022online,moon2023online} and repetition of previously seen concepts \citep{hemati2024continual} have been used. Benchmarks for lifelong learning have also been constructed based on data from autonomous driving \citep{verwimp2023clad}, using image datasets collected through time \citep{bornschein2023nevis} and by concatenating thousands of short videos \citep{carreira2024learning}.

\section{Datasets}
\label{sec:datasets}

Exploring the possibility of learning video models in a lifelong fashion requires (1) long continuous video streams
and (2) varying levels of complexity to gauge challenges and limitations.
As discussed in \Cref{sec:background},
existing video datasets usually consist of many short videos that may or may not be temporally related.
To that end, we introduce five new video datasets derived from single video streams without semantic discontinuities that vary in perceptual complexity, stochasticity, temporal correlation, and degree of nonstationarity (refer to \Cref{appendix:curation} for more details).



\paragraph{Lifelong Bouncing Balls (O)} contains 1 million 32x32 RGB video frames for training ($\sim$28 hours long at 10FPS) and another 1 million frames for evaluation.
The video contains two colored balls that deterministically bounce around in a 2D environment, colliding with boundaries and each other.
Upon collision, each ball changes velocity according to the conservation of momentum and cycles through a repeating color sequence of red, yellow, red, and green.
Since the ball states are fully observable and the transition dynamics are deterministic, the future frames can be perfectly predicted.
Solving \textit{Lifelong Bouncing Balls (O)} requires learning the 2D environment's \textbf{deterministic} dynamics and retaining the balls' color transition histories from a \textbf{correlated} and \textbf{repetitive} video stream.

\paragraph{Lifelong Bouncing Balls (C)} introduces nonstationarity on top of \textit{Lifelong Bouncing Balls (O)}.
While the balls' motion matches that of (O),
the blue channel values of all ball colors increase over time at a constant rate (red, yellow, and green ball colors respectively become fuchsia, white, and aqua by the end of the video).
The evaluation set contains frames with balls of all previously observed colors. 
Solving \textit{Lifelong Bouncing Balls (C)} requires learning the \textbf{deterministic} ball dynamics and color transitions from a \textbf{correlated} video stream with \textbf{unrepetitive} details.

\paragraph{Lifelong 3D Maze}
contains 1 million 64x64 RGB video frames for training ($\sim$14 hours long at 20FPS) and 100,000 frames for evaluation.
The video is a first-person view of an agent that navigates a randomly generated 3D maze.
Whenever the agent solves a maze, the walls come down and rise again to form a new, unseen maze.
The mazes contain various sparse objects, including rocks that flip the agent upside down on contact.
Since maze states are partially observable and transitions are stochastic, future frames cannot be perfectly predicted from past ones.
Solving \textit{Lifelong 3D Maze} requires learning the first-person sensory inputs associated with navigating a \textbf{stochastic} 3D environment and modeling \textbf{infrequent events} from a largely \textbf{repetitive} and \textbf{correlated} video stream.

\paragraph{Lifelong Drive}
contains 1 million 512x512 RGB video frames for training ($\sim$14 hours long at 20FPS) and 100,000 frames for evaluation.
We encode these frames using the Stable Diffusion perceptual encoder \citep{rombach2022high} to 4x64x64 frames.
The video is continuous dashcam footage of a 1675km drive from Chongqing to Shanghai \citep{china2023kilometers}.
The drive spans diverse scenery such as mountains, cities, and tunnels, weather conditions such as sunny, cloudy, and rainy weather, and other cars and trucks moving at different speeds.
Solving \textit{Lifelong Drive} requires modeling \textbf{real world perceptual data} from a single \textbf{correlated} and \textbf{nonstationary} video stream. 

\paragraph{Lifelong PLAICraft}
contains 1 million 1280x768 RGB video frames for training ($\sim$28 hours long at 10FPS) and 500,000 frames for evaluation.
We encode these frames using the Stable Diffusion perceptual encoder to 4x160x96 frames.
The training video is a first-person view of a player with an ID of ``Alex'' in a multiplayer Minecraft survival world \citep{he2025plaicraft}.
The evaluation video is a first-person view of another player with an ID of ``Kyrie'' who explores the same world.
The videos span months of continuous gameplay sessions, covering diverse dimensions, biomes, mining, crafting, construction, mob fighting, and player-to-player interactions.
The Minecraft world contains aspects that repeat (ex. day-night cycle, players visiting their homes) and do not repeat (ex. felled trees, player chat logs).
Solving \textit{Lifelong PLAICraft} requires learning a \textbf{highly nonstationary} environment that changes in \textbf{multiple timescales} from a single \textbf{correlated} video stream.

\begin{figure*}[t!]
    \centering    \includegraphics[width=0.875\linewidth]{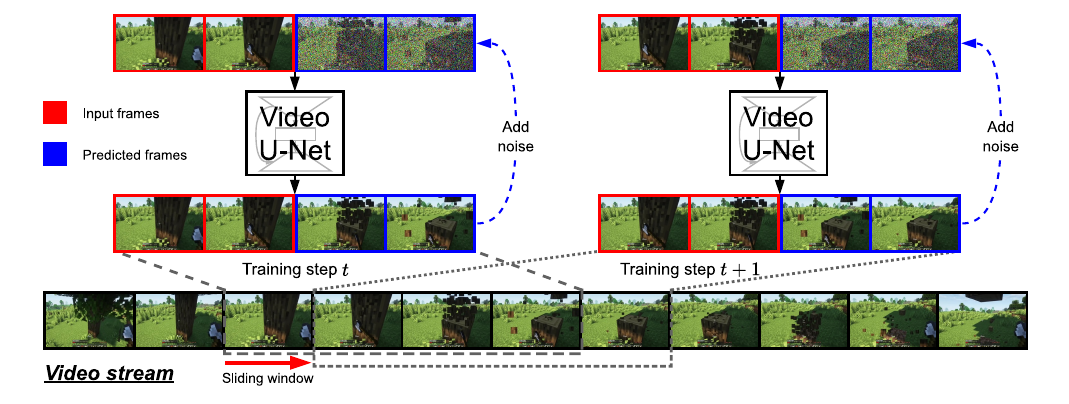}
    \caption{Video diffusion model lifelong learning on a single video stream with $K\!=\!4$. At training step $t$, the model conditions on the frames in the first half of its context window (red) and learns to denoise the frames in the second half of its context window (blue). At training step $t+1$, the model's context window shifts by one video frame, and the same procedure repeats indefinitely.}
    \label{fig:diagram}
\end{figure*}

\section{Model and Training Regime}
\label{sec:methodology}

\paragraph{Model.}
We use U-Net \citep{harvey2022flexible} and Transformer \citep{lu2024vdt} video diffusion models.
The networks operate on $K$ frames at a time,
where $K=10$ for \textit{Lifelong Bouncing Balls} and \textit{Lifelong PLAICraft} and $K=20$ otherwise.
The first $K/2$ frames are model inputs and the next $K/2$ frames are prediction targets.

During training, given the first $K/2$ uncorrupted video frames and the second $K/2$ video frames corrupted with Gaussian noise, the model regresses the values of the Gaussian noise \citep{ho2020denoising}.
Specifically, let $\boldsymbol{x}$ be a size $K$ window of consecutive video frames, and $F_{\boldsymbol\theta}$ be the denoising neural network.
The denoising loss function is defined as
\begin{equation}
    \ell(\boldsymbol\theta, \boldsymbol{x}) = \mathbb{E}_{\boldsymbol{\epsilon},s} \left\| \boldsymbol{\epsilon} - F_{\boldsymbol\theta}(\boldsymbol{x}^{\mathrm{obs}}, \boldsymbol{x}^{\mathrm{lat}}_s, s) \right\|^2,
    \label{eq:denoising-loss}
\end{equation}%
where $\epsilon \sim \mathcal{N}(0,I)$ is a unit normal random vector, and $s \sim \mathcal{U}(1, S)$ is the diffusion noising timestep. The superscripts $\mathrm{obs}$ and $\mathrm{lat}$ respectively represent the input and predicted part of $\boldsymbol{x}$ i.e., the first and second $K/2$ frames. The subscript $s$ denotes that the video frames are corrupted to the $s$-th diffusion timestep using $\epsilon$ as noise \citep{ho2020denoising}.
During sampling, given the first $K/2$ clean video frames $\boldsymbol{x}^{\mathrm{obs}}$ and the second $K/2$ video frames $\boldsymbol{x}^{\mathrm{lat}}$ filled with Gaussian noise, the model iteratively denoises $\boldsymbol{x}^{\mathrm{lat}}$ to produce a plausible continuation of $\boldsymbol{x}^{\mathrm{obs}}$ using the stochastic sampler of Karras et al.~\cite{karras2022elucidating}.
For the rest of this section, $\mathcal{X}$ denotes the video stream, $\mathcal{X}^{i:i+K}$ denotes a length-$K$ sliding window starting at frame $i$, and \Cref{eq:denoising-loss}'s expectation is approximated with a single-sample estimation.

\paragraph{Baseline: Offline (i.i.d.) Learning.}
As a baseline representative of standard video models, we train models in an \emph{Offline Learning} regime.
The $t$-th training step loss is $\mathcal{L}^{\mathrm{offline}}_t(\boldsymbol{\theta}, \mathcal{X}) = \ell(\boldsymbol{\theta}, \mathcal{X}^{i:i+K})$,
where $i$ is a uniformly sampled video frame index.
The resulting training data resembles i.i.d.\ segments of $K$ video frames, even though all frames originate from a single autocorrelated stream.
When using batch size $N > 1$ (as is common), this i.i.d.\ sampling is repeated $N$ times.

\paragraph{Lifelong Learning.}
While the Offline Learning regime randomly shuffles the data during training,
the Lifelong Learning regime presents data to the model \emph{in order}.
This setup reflects the real-time streaming learning setting in \Cref{sec:intro}.
The $t$-th training step loss is $\mathcal{L}^{\mathrm{lifelong}}_t(\boldsymbol{\theta}, \mathcal{X}) = \ell(\boldsymbol{\theta}, \mathcal{X}^{t:t+K})$,
i.e., the model receives a sliding window from the single video $\mathcal{X}$.
When using batch size $N>1$, we may fill the batch with multiple copies of the current video $\mathcal{X}^{t:t+K}$.

This Lifelong Learning setup is compatible with various techniques from the continual learning community.
As a demonstration, we augment the online stream with a replay buffer \citep{chaudhry2019tiny}
that retains a buffer $\mathcal{M}$ of past video subsequences chosen through reservoir sampling \citep{vitter1985reservoir},
with a fixed size replay buffer.
With a batch size $N$, $C_t$ copies of the latest subsequence in the batch, and replay samples $\mathcal{M}^{i}$ drawn from the buffer, the new $t$-th training step loss is
\begin{equation}
    \mathcal{L}^{\mathrm{lifelong}}_t(\boldsymbol{\theta}, \mathcal{X}) =
    \tfrac{1}{N}\!\!\left(\!C_t\,\ell(\boldsymbol{\theta}, \mathcal{X}^{t:t+K}\!)
     + \!\!\textstyle\sum_{i=1}^{N-C_t}\!\ell(\boldsymbol{\theta}, \mathcal{M}^{i})\!\right)\!.
     \label{eq:replay}
\end{equation}%
We refer to this algorithm as \textit{Lifelong Learning} for the rest of the paper.
Optimizing $\mathcal{L}^{\mathrm{lifelong}}_t$ at each timestep, even with an infinite-size buffer, differs from optimizing $\mathcal{L}^{\mathrm{offline}}_t$ because the former only has access to frames up to time $t$ and always includes the most recent frames in the loss.

\section{Experiments}
\label{sec:experiments}

This section qualitatively and quantitatively compares offline learned and lifelong learned video diffusion models.
The learning algorithms
use the same batch size and number of gradient steps, equalizing the amount of computation and runtime memory.
We also evaluate two additional baselines that update model parameters using only the most recent $K$ frames, without replay buffers: Streaming AdamW, based on standard AdamW \citep{loshchilov2018decoupled}, and Orthogonal AdamW, based on the optimizer of Han et al. \citep{han2025learning} designed for discriminative model lifelong learning.
We refer the readers to \Cref{appendix:expdetails} for additional experiment details. 

\subsection{Lifelong Bouncing Balls}
\label{subsec/ball_experiment}

\paragraph{Setup.}
As discussed in \Cref{sec:datasets}, the 28 hour long \textit{Lifelong Bouncing Balls} datasets, with stationary (Version O) and non-stationary (Version C) color transitions,  feature deterministic dynamics that are Markovian on the time-scale of two ball bounces; intentionally longer than the context window of our model.
The U-Net diffusion models for \textit{Lifelong Bouncing Balls (O)} and \textit{Lifelong Bouncing Balls (C)} respectively have 8 and 74 million parameters, whereas the Transformer diffusion models for both datasets have 42 million parameters.
All models are trained with a batch size of 2 and are evaluated using 45 frames that they autoregressively generate conditioned on 5 ground truth frames.
Lifelong Learning uses a 1.5 hour replay buffer.
We measure the sampled videos' perceptual and temporal coherence using FVD \citep{unterthiner2019accurate} and report the loss to measure learning progress.
We also measure the models' understanding of the ball movement using \mbox{minADE} \citep{rasouli2020deep} and the color transition using our novel ColorKL metric (\Cref{appendix:metrics}).

\begin{figure}[t]
    \centering
    \begin{minipage}[t]{0.475\linewidth}
        \begin{subfigure}{\linewidth}
        \centering
        \includegraphics[width=5.5cm, height=3.5cm, keepaspectratio=false]{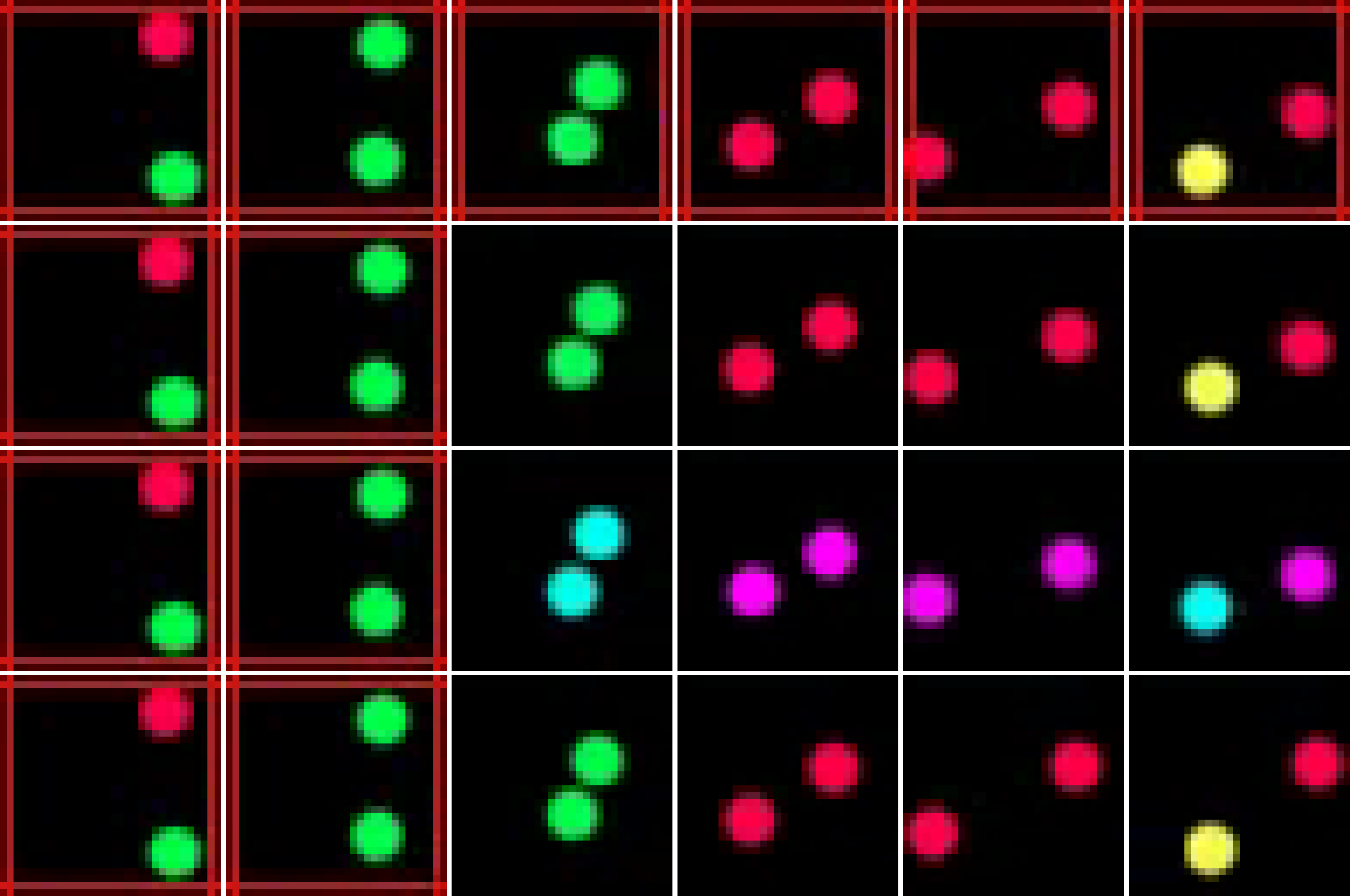}
        \caption{\textit{Lifelong Bouncing Balls (C)} videos.}
        \label{fig:forgetting-ball-nstn}
        \end{subfigure}
    \end{minipage}
    \hfill
    \begin{minipage}[t]{0.475\linewidth}
        \begin{subfigure}{\linewidth}
        \centering
        \includegraphics[width=5.5cm, height=3.5cm, keepaspectratio=false]{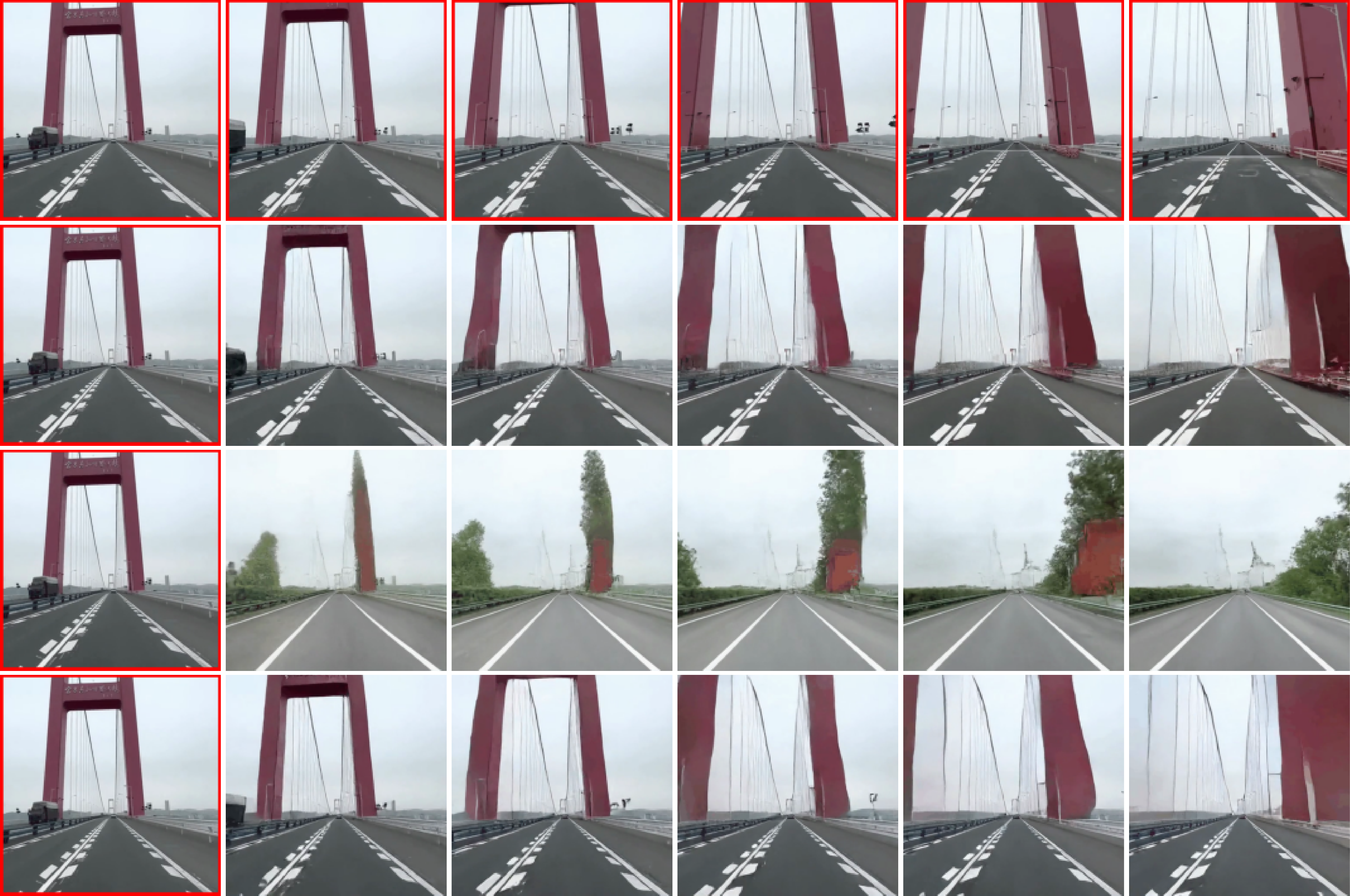}
        \caption{\textit{Lifelong Drive} videos.}
        \label{fig:forgetting-drive}
        \end{subfigure}
    \end{minipage}
    \caption{Example results for nonstationary datasets.
    In each panel, rows one through four show ground-truth videos and samples from models trained with Offline Learning, Streaming AdamW, and Lifelong Learning, respectively.
    While Offline Learning and Lifelong Learning samples more faithfully continue the conditioning frames highlighted in \red{red}, Streaming AdamW samples lose details not readily available in the recent training sliding windows (e.g., ball colors and bridge structure).}
\end{figure}

{%
\captionsetup[table]{position=top,skip=4pt}
\begin{table*}[t!]
    \centering
    \caption{\textit{Lifelong Bouncing Balls} metrics ($\downarrow$) computed from 2 training and 3 sampling seeds. The labels (U) and (T) denote U-Net and Transformer architectures.}
    \ra{1.05}
    \small
    \setlength{\tabcolsep}{3pt}
    \begin{subtable}{\textwidth}
    \centering
    \caption{\textit{Lifelong Bouncing Balls (O)}}
    \label{tab:ball_stn}
    \begin{adjustbox}{width=\textwidth}
    \begin{tabular}{@{}*{9}{c}@{}}
        \toprule
        & \multicolumn{4}{c}{\textit{Train Stream}} & \multicolumn{4}{c}{\textit{Test Stream}}\\
        \cmidrule(lr){2-5} \cmidrule(lr){6-9}

        Method & FVD & Loss$^{\times \! 10^{-\!5}}$ & minADE  & ColorKL  & FVD  & Loss$^{\times \! 10^{-\!5}}$ & minADE  & ColorKL  \\
        \midrule

Offline Learning (U) & 4.5 \tiny{±0.1} & 6.3 \tiny{±0.1} & 1.74 \tiny{±0.03} & 0.006 \tiny{±0.001} & 4.7 \tiny{±0.2} & 6.2 \tiny{±0.2} & 1.71 \tiny{±0.06} & 0.006 \tiny{±0.001} \\

Lifelong Learning (U) & 4.9 \tiny{±0.2} & 6.3 \tiny{±0.2} & 1.82 \tiny{±0.03} & 0.005 \tiny{±0.000} & 4.7 \tiny{±0.2} & 6.2 \tiny{±0.3} & 1.81 \tiny{±0.03} & 0.005 \tiny{±0.000} \\

Streaming AdamW (U) & 4.7 \tiny{±0.1} & 6.7 \tiny{±0.1} & 1.91 \tiny{±0.01} & 0.006 \tiny{±0.0} & 5.0 \tiny{±0.2} & 6.7 \tiny{±0.1} & 1.84 \tiny{±0.03} & 0.005 \tiny{±0.0} \\

Orthogonal AdamW (U) & 5.2 \tiny{±0.2} & 6.8 \tiny{±0.1} & 1.99 \tiny{±0.09} & 0.005 \tiny{±0.001} & 4.8 \tiny{±0.1} & 6.7 \tiny{±0.1} & 1.92 \tiny{±0.07} & 0.005 \tiny{±0.0} \\

\midrule

Offline Learning (T) & 5.4 \tiny{±0.2} & 6.8 \tiny{±0.3} & 2.25 \tiny{±0.12} & 0.006 \tiny{±0.001} & 5.1 \tiny{±0.2} & 6.7 \tiny{±0.3} & 2.16 \tiny{±0.12} & 0.006 \tiny{±0.001} \\

Lifelong Learning (T) & 5.5 \tiny{±0.1} & 6.6 \tiny{±0.2} & 2.22 \tiny{±0.05} & 0.005 \tiny{±0.001} & 5.0 \tiny{±0.3} & 6.5 \tiny{±0.2} & 2.16 \tiny{±0.06} & 0.005 \tiny{±0.001} \\

        \bottomrule
    \end{tabular}
    \end{adjustbox}
\end{subtable}
\begin{subtable}{\textwidth}
    \centering
    \caption{\textit{Lifelong Bouncing Balls (C)}}
    \label{tab:ball_nstn}
    \ra{1.05}
    \small
    \setlength{\tabcolsep}{3pt}
    \begin{adjustbox}{width=\textwidth}
    \begin{tabular}{@{}*{9}{c}@{}}
        \toprule
        & \multicolumn{4}{c}{\textit{Train Stream}} & \multicolumn{4}{c}{\textit{Test Stream}}\\
        \cmidrule(lr){2-5} \cmidrule(lr){6-9}
        Method & FVD  & Loss$^{\times \! 10^{-\!5}}$ & minADE  & ColorKL  & FVD  & Loss$^{\times \! 10^{-\!5}}$ & minADE  & ColorKL  \\
        \midrule

Offline Learning (U) & 5.8 \tiny{±0.3} & 6.5 \tiny{±0.1} & 2.04 \tiny{±0.09} & 0.007 \tiny{±0.002} & 5.9 \tiny{±0.2} & 6.5 \tiny{±0.1} & 2.14 \tiny{±0.10} & 0.007 \tiny{±0.001} \\

Lifelong Learning (U) & 5.0 \tiny{±0.1} & 7.4 \tiny{±0.1} & 2.03 \tiny{±0.00} & 0.005 \tiny{±0.001} & 5.7 \tiny{±0.2} & 7.5 \tiny{±0.1} & 2.06 \tiny{±0.00} & 0.005 \tiny{±0.000} \\

Streaming AdamW (U) & 357.4 \tiny{±1.8} & 2240 \tiny{±110} & 2.61 \tiny{±0.08} & 0.021 \tiny{±0.002} & 343.6 \tiny{±1.2} & 2252 \tiny{±108} & 2.73 \tiny{±0.11} & 0.022 \tiny{±0.0} \\

Orthogonal AdamW (U) & 355.0 \tiny{±1.7} & 2327 \tiny{±13} & 2.72 \tiny{±0.02} & 0.024 \tiny{±0.001} & 345.4 \tiny{±1.3} & 2301 \tiny{±13} & 2.83 \tiny{±0.01} & 0.023 \tiny{±0.003} \\

\midrule

Offline Learning (T) & 5.0 \tiny{±0.2} & 8.4 \tiny{±0.1} & 2.24 \tiny{±0.05} & 0.006 \tiny{±0.000} & 5.6 \tiny{±0.1} & 8.5 \tiny{±0.1} & 2.19 \tiny{±0.02} & 0.006 \tiny{±0.000} \\

Lifelong Learning (T) & 5.3 \tiny{±0.1} & 8.1 \tiny{±0.0} & 2.29 \tiny{±0.03} & 0.006 \tiny{±0.001} & 6.4 \tiny{±0.3} & 8.1 \tiny{±0.0} & 2.27 \tiny{±0.03} & 0.006 \tiny{±0.000} \\

        \bottomrule
    \end{tabular}
    \end{adjustbox}
    \end{subtable}
\end{table*}

}

\paragraph{Result.}

Qualitative results appear in \Cref{fig:ball_examples} and \Cref{appendix:additional_samples}.
For both \textit{Lifelong Bouncing Balls (O)} and \textit{Lifelong Bouncing Balls (C)}, Offline Learning and Lifelong Learning produce models that learn the longer than context window Markov-order ball color and Newtonian dynamics.
The videos generated by offline and lifelong learned models are not perceptibly different in quality.
Even under the nonstationary color changes in \textit{Lifelong Bouncing Balls (C)}, both methods produce realistic continuations for frames from both the beginning and end of the stream.
The quantitative results in \Cref{tab:ball_stn} and \Cref{tab:ball_nstn} mirror these findings.
However, Streaming AdamW and Orthogonal AdamW perform comparably to the aforementioned methods on \textit{Lifelong Bouncing Balls (O)} but significantly underperform on \textit{Lifelong Bouncing Balls (C)} from forgetting the earlier ball colors (see \Cref{fig:forgetting-ball-nstn}).
This suggests that while short-range temporal correlation can be learned by optimizer-based streaming  methods, techniques that mitigate forgetting are required to address non-stationarity.
Overall, temporal correlation and the degree of non-repetitiveness do not pose a significant challenge to Lifelong Learning with replay buffers.

\subsection{Lifelong 3D Maze}
\label{subsec/maze_experiment}

\paragraph{Setup.}
The 14 hour long \emph{Lifelong 3D Maze} dataset is a step up in complexity from \emph{Lifelong Bouncing Balls}.
The video stream frames are renderings of a 3D space and feature stochastic dynamics and rare events.
The U-Net and Transformer-based diffusion models from this section respectively have 78 million and 110 million parameters.
All models are trained with a batch size of 4, of which two elements are the current timestep video frame subsequence, and are evaluated using 40 frames autoregressively generated from conditioning on 10 ground truth frames.
Lifelong Learning uses a 42 minute replay buffer.
We measure perceptual and temporal coherence using FVD and JEDi \citep{luo2025beyond} and learning progress using the diffusion loss.

\paragraph{Result.}

Qualitative results appear in \Cref{fig:maze_examples} and \Cref{appendix:additional_samples}.
Both Offline Learning and Lifelong Learning models generate coherent maze trajectories,
successfully modeling sparsely occurring events such as camera inversion and the rising of the maze walls.
Both models make occasional mistakes, such as deforming rocks during camera inversion.
Quantitative results appear in \Cref{tab:maze}.
We find that the Offline and Lifelong Learning metrics are similar, and neither methods significantly outperform the other across all metrics, architectures, and stream types.
Streaming AdamW and Orthogonal AdamW perform comparably worse, with worse train stream performance than test stream performance due to forgetting from the former’s mild distribution shift (see \Cref{appendix:curation}).
Overall, we see on qualitative and quantitative fronts that Lifelong Learning performs well and can capture rare event information when learning from correlated frames from a stochastic environment, despite heavy data imbalance between rare events and regular maze traversal frames.

\begin{table*}[t!]
    \centering
    \caption{\textit{Lifelong 3D Maze} metrics ($\downarrow$) computed from 1 training and 3 sampling seeds.}
    \label{tab:maze}
    \ra{1.05}
    \small
    \setlength{\tabcolsep}{4.5pt}
    \begin{adjustbox}{width=\textwidth}
    \begin{tabular}{@{}*{7}{c}@{}}
        \toprule
        & \multicolumn{3}{c}{\textit{Train Stream}} & \multicolumn{3}{c}{\textit{Test Stream}}\\
        \cmidrule(lr){2-4} \cmidrule(lr){5-7}
        Method & FVD  & JEDi  & Loss  & FVD  & JEDi  & Loss \\
        \midrule
    
Offline Learning (U-Net) & 34.2 \tiny{±1.3} & 0.083 \tiny{±0.002} & 0.005 \tiny{±0.0} & 28.5 \tiny{±1.2} & 0.060 \tiny{±0.001} & 0.005 \tiny{±0.0} \\

Lifelong Learning (U-Net) & 41.2 \tiny{±0.4} & 0.073 \tiny{±0.006} & 0.006 \tiny{±0.0} & 30.7 \tiny{±0.1} & 0.061 \tiny{±0.003} & 0.006 \tiny{±0.0} \\

Streaming AdamW (U-Net) & 130.6 \tiny{±0.6} & 0.142 \tiny{±0.002} & 0.009 \tiny{±0.0} & 36.9 \tiny{±1.9} & 0.104 \tiny{±0.004} & 0.006 \tiny{±0.0} \\

Orthogonal AdamW (U-Net) & 135.0 \tiny{±2.1} & 0.140 \tiny{±0.002} & 0.009 \tiny{±0.0} & 45.9 \tiny{±0.4} & 0.129 \tiny{±0.004} & 0.006 \tiny{±0.0} \\

        \midrule
Offline Learning (Transformer) & 30.2 \tiny{±0.9} & 0.088 \tiny{±0.005} & 0.006 \tiny{±0.0} & 32.6 \tiny{±1.0} & 0.082 \tiny{±0.005} & 0.006 \tiny{±0.0} \\

Lifelong Learning (Transformer) & 32.1 \tiny{±0.9} & 0.082 \tiny{±0.005} & 0.006 \tiny{±0.0} & 30.2 \tiny{±0.8} & 0.070 \tiny{±0.001} & 0.006 \tiny{±0.0} \\

        \bottomrule
    \end{tabular}
    \end{adjustbox}
\end{table*}

\begin{table*}[t!]
    \centering
    \caption{\textit{Lifelong Drive} metrics ($\downarrow$) computed from 1 training and 3 sampling seeds.}
    \label{tab:drive}
    \setlength{\tabcolsep}{4.5pt}
    \ra{1.05}
    \small
    \centering
    \begin{adjustbox}{width=\textwidth}
    \begin{tabular}{@{}*{7}{c}@{}}
        \toprule
        & \multicolumn{3}{c}{\textit{Train Stream}} & \multicolumn{3}{c}{\textit{Test Stream}}\\
        \cmidrule(lr){2-4} \cmidrule(lr){5-7}
        Method & FVD  & JEDi  & Loss 
        & FVD  & JEDi  & Loss  \\
        \midrule

Offline Learning (U-Net) & 15.3 \tiny{±0.2} & 0.071 \tiny{±0.000} & 0.029 \tiny{±0.0} & 17.9 \tiny{±0.3} & 0.102 \tiny{±0.001} & 0.026 \tiny{±0.0} \\

Lifelong Learning (U-Net) & 17.0 \tiny{±0.2} & 0.072 \tiny{±0.000} & 0.030 \tiny{±0.0} & 21.8 \tiny{±0.3} & 0.106 \tiny{±0.000} & 0.027 \tiny{±0.0} \\

Streaming AdamW (U-Net) & 322.8 \tiny{±1.2} & 0.608 \tiny{±0.002} & 0.041 \tiny{±0.0} & 183.4 \tiny{±1.7} & 0.343 \tiny{±0.000} & 0.029 \tiny{±0.0} \\

Orthogonal AdamW (U-Net) & 294.4 \tiny{±2.3} & 0.590 \tiny{±0.001} & 0.041 \tiny{±0.0} & 180.0 \tiny{±0.7} & 0.315 \tiny{±0.0} & 0.029 \tiny{±0.0} \\

        \midrule

Offline Learning (Transformer) & 10.4 \tiny{±0.2} & 0.023 \tiny{±0.000} & 0.027 \tiny{±0.0} & 11.2 \tiny{±0.1} & 0.036 \tiny{±0.000} & 0.024 \tiny{±0.0} \\

Lifelong Learning (Transformer) & 10.7 \tiny{±0.1} & 0.023 \tiny{±0.000} & 0.029 \tiny{±0.0} & 12.9 \tiny{±0.1} & 0.039 \tiny{±0.000} & 0.025 \tiny{±0.0} \\

        \bottomrule
    \end{tabular}
    \end{adjustbox}
\end{table*}

\subsection{Lifelong Drive}
\label{subsec/drive_experiment}

\paragraph{Setup.}
The 14 hour long \textit{Lifelong Drive} dataset features real-world perceptual data captured in a 3D environment similar to \textit{Lifelong 3D Maze}, but with higher frame resolution and greater perceptual complexity.
The U-Net and Transformer-based diffusion models in this section respectively have 80 and 300 million parameters.
Lifelong Learning uses a 2.8 hour replay buffer (42 minute results in \Cref{tab:drive_full}).
All models are trained with a batch size of 8, of which two elements are the current timestep video frame subsequence, and evaluation is performed on videos with 20 frames, 10 of which are model-generated given the preceding 10 frames.
We report FVD, JEDi, and model loss.

\paragraph{Result.}

See \Cref{fig:drive_examples} and \Cref{appendix:additional_samples} for qualitative results. 
Similar to 3D Maze, Lifelong Learning models produce realistic 3D navigation dynamics.
The generated videos from models trained with either algorithm are realistic and capture actions like lane changes,
though a mild autoregressive drift appears that is less noticeable in simpler datasets (e.g. \Cref{fig:forgetting-drive}'s bridge shape).
Quantitative results appear in \Cref{tab:drive}.
Offline and Lifelong Learning perform similarly across metrics, architectures, and stream types.
We see that changing the model architecture and size affects performance far more than the choice of learning algorithm, suggesting that video diffusion models can be effectively lifelong learned from an autocorrelated video stream.
Streaming AdamW and Orthogonal AdamW results suggest that learning from an in-order stream without replay buffers exhibits catastrophic forgetting of early colors (Balls) and structures (Drive) \citep{robins1995catastrophic} but interestingly retains scene dynamics understanding (see \Cref{fig:forgetting-drive}).

\subsection{Lifelong PLAICraft}
\label{subsec/minecraft_experiment}

\begin{table*}[t!]
    \centering
    \caption{\textit{Lifelong PLAICraft} metrics ($\downarrow$) computed from 1 training and 3 sampling seeds.}
    \label{tab:minecraft}
    \ra{1.05}
    \small
    \setlength{\tabcolsep}{4.5pt}
    \begin{adjustbox}{width=\textwidth}
    \begin{tabular}{@{}*{7}{c}@{}}
        \toprule
        & \multicolumn{3}{c}{\textit{Train Stream}} & \multicolumn{3}{c}{\textit{Test Stream}}\\
        \cmidrule(lr){2-4} \cmidrule(lr){5-7}
        Method & FVD  & JEDi  & Loss 
        & FVD  & JEDi  & Loss  \\
        \midrule

Offline Learning (U-Net) & 59.7 \tiny{±0.2} & 0.874 \tiny{±0.007} & 0.034 \tiny{±0.0} & 119.8 \tiny{±1.3} & 1.119 \tiny{±0.004} & 0.042 \tiny{±0.0} \\

Lifelong Learning (U-Net) & 62.9 \tiny{±0.6} & 0.876 \tiny{±0.003} & 0.034 \tiny{±0.0} & 130.8 \tiny{±1.1} & 1.189 \tiny{±0.003} & 0.042 \tiny{±0.0} \\

Streaming AdamW (U-Net) & 239.5 \tiny{±1.5} & 2.038 \tiny{±0.014} & 0.047 \tiny{±0.0} & 270.6 \tiny{±1.9} & 2.281 \tiny{±0.004} & 0.050 \tiny{±0.0} \\

Orthogonal AdamW (U-Net) & 220.5 \tiny{±0.8} & 1.608 \tiny{±0.006} & 0.048 \tiny{±0.0} & 251.7 \tiny{±0.8} & 1.824 \tiny{±0.009} & 0.052 \tiny{±0.0} \\

        \midrule

Offline Learning (Transformer) & 23.7 \tiny{±0.1} & 0.209 \tiny{±0.001} & 0.024 \tiny{±0.0} & 27.7 \tiny{±0.3} & 0.223 \tiny{±0.002} & 0.024 \tiny{±0.0} \\

Lifelong Learning (Transformer) & 21.9 \tiny{±0.4} & 0.197 \tiny{±0.003} & 0.025 \tiny{±0.0} & 26.3 \tiny{±0.3} & 0.208 \tiny{±0.001} & 0.024 \tiny{±0.0} \\

        \bottomrule
    \end{tabular}
    \end{adjustbox}
\end{table*}

\paragraph{Setup.}
The 28 hour long \emph{Lifelong PLAICraft} dataset is the most challenging in terms of temporal dynamics out of the five datasets.
The video stream is continuously recorded from a partially observable environment containing multi-agent interaction and a highly diverse perceptual and action space.
The U-Net and Transformer-based diffusion models in this section respectively have 80 and 300 million parameters.
Lifelong Learning uses a 5.6 hour replay buffer (1.4 hour results in \Cref{tab:minecraft_full}).
All models are trained with a batch size of 8, of which two elements are the current timestep video frame subsequence, and evaluation is performed on videos with 10 frames, 5 of which are model-generated conditioned on the preceding 5 frames.
We report FVD, JEDi, and model loss.

\paragraph{Result.}
See \Cref{fig:minecraft_examples} and \Cref{appendix:additional_samples} for qualitative results. 
Both Offline and Lifelong Learning models capture perceptual details about Minecraft videos (e.g., player name, equipped item, chat logs in \Cref{fig:minecraft_examples}).
Frequently observed temporal dynamics (e.g. item movement when walking and flashing red dots), are captured better than player movement and actions.
We hypothesize that larger models and hyperparameter tuning on commercial-scale compute will significantly improve sample quality for both algorithms.
Nevertheless, the generated videos from offline and lifelong learned models are not perceptually distinguishable in quality.
Quantitative results in \Cref{tab:minecraft} confirm this: on both training and test streams, neither approach consistently outperforms the other across metrics and architectures, while Streaming AdamW and Orthogonal AdamW underperform Lifelong Learning.
In summary, we find that Lifelong Learning can perform comparably to Offline Learning with minimal hyperparameter tuning, even on videos as complex as \emph{Lifelong PLAICraft}.

\subsection{Discussion}

The metrics for Offline and Lifelong Learning are largely similar across datasets.
While the standard errors for the two learning algorithms do not always overlap for every metric, dataset, and model architecture, these differences are small relative to the effects of control variables such as model architecture, parameter size, dataset, and optimizer choice (see \Cref{appendix:orthogonal_replay} results on optimizing the replay loss with Orthogonal AdamW).
We additionally observe in \Cref{appendix:timeline} that Lifelong Learning performs similarly to Offline Learning on video subsequences from different training timesteps, further reinforcing the claim that forgetting arising from nonstationarity is not a critical obstacle given replay buffers.
Combined with the qualitative indistinguishability of offline and lifelong model samples, we conclude that video diffusion models can be effectively learned in a lifelong manner by overcoming data stream autocorrelation and nonstationarity.

\paragraph{Replay Buffer Size Ablation Studies.}
We investigate model sensitivity to replay buffer size in \Cref{appendix:quantitative}.
We find that using a replay buffer four times smaller than above results in a slight performance drop that is much smaller in relative terms than using Streaming AdamW or Orthogonal AdamW (\Cref{tab:drive_full}, \Cref{tab:minecraft_full}). 
We find that replay buffers are necessary but that increasing replay buffer sizes beyond those above did not given rise to significantly different performance. 

\paragraph{Human Evaluation.}
We conduct human evaluation on 76 Offline versus Lifelong Learning clip pairs illustrating key dataset characteristics (e.g., screen flip rare event for Lifelong 3D Maze, lane change for Lifelong Drive) in \Cref{appendix:human_eval}.
For each pair, five Amazon Mechanical Turk participants indicated—without knowing which model produced which clip—whether one was of higher quality or the two appeared similar.
Across datasets, participants show no consistent preference for either method, even when excluding responses marked ``similar,'' indicating that the perceptual quality of lifelong-trained models is on par with their offline counterparts.


\paragraph{Continued Learning.}
For the U-Net models, we observe that test performance continues to improve as more data is observed for all algorithms (\Cref{appendix:online_test}). 
Having controlled for equal training computation throughout, we now relax that assumption and ask how many more frames the slightly less performant Minecraft continual learning model needs to observe to exceed the test performance of the offline learned model.
We continued training the Lifelong Learning Minecraft U-Net model on 22.4 more hours and found that doing so yielded a model closer to or even exceeding the offline learned model on test performance (118.0 {\small ±0.5} FVD, 1.150 {\small ±0.002} JEDi, 0.041 {\small ±0.0} Loss).
This suggests that even in situations where Offline Learning achieves slightly higher model performance for equal compute, lifelong learned video diffusion learning can achieve comparable performance with simply more data in a natural, streaming fashion.

\paragraph{EgoLife Results.}
To test whether our findings extend to datasets constructed from multiple long videos, we additionally assess Offline and Lifelong Learning on one million frames associated with the \emph{A1\_JAKE} participant from the EgoLife dataset \citep{yang2025egolife}. 
The EgoLife video stream comprises ten concatenated video clips depicting daily activities and human interactions across indoor and outdoor environments, and therefore deviates from our single, continuous video stream setting.
As shown in \Cref{appendix:egolife}, Lifelong Learning performance remains generally similar to that of Offline Learning, where the choice of architecture produces substantially larger performance differences than the choice of training regime.
Overall, these results further support the generality of our main findings.
\section{Related Work}
\label{sec:relatedwork}

Our work is closely related to continual generative modeling \citep{nguyen2018variational,ramapuram2020lifelong,masip2023continual,smith2024continual,zajac2023exploring,campo2020continual,chen2022continual}, in particular diffusion and video generative models. Prior work on diffusion model lifelong learning \citep{masip2023continual,smith2024continual,zajac2023exploring,liu2025can} focuses on image modeling in task-based continual learning. Prior work on video generative model lifelong learning \citep{campo2020continual,chen2022continual} likewise trains VAE models that generate future frames in the task-based continual learning setting. Unlike our setup, these work assume rigid task boundaries in data streams and the ability to train to convergence on large data stream batches. In contrast, our lifelong learning datasets do not have notions of tasks, and our models only access data as they appear in the stream, a setup that better reflects biological and embodied-agent learning streams.

Our work is also related to online learning of sequence processing models \citep{zucchet2023online, carreira2024learning, bornschein2024transformers, liu2024locality, han2025learning, elsayed2025sequence}. Notably, Carreira et al.~\cite{carreira2024learning} and Han et al.~\cite{han2025learning} learn discriminative video prediction models from the concatenation of loosely related short to medium-length videos, whereas we learn generative video models from a single autocorrelated stream. Liu et al.~\cite{liu2024locality} learn an online linear-regression world model, which, as they note, is not amenable to high-dimensional videos.

Lastly, prior work has, like us, introduced video datasets for continual learning.
The datasets introduced by Villa et al.~\cite{villa2022vclimb} and Tang et al.~\cite{tang2024vilco} contain many short videos that can be learned under a task-based continual learning setup. Carreira et al.~\cite{carreira2024learning} construct one very long data stream by concatenating multiple short-to-medium-length videos, but their data are not publicly available. The datasets of Singh et al.~\cite{singh2016krishnacam} and Yang et al.~\cite{yang2025egolife}, while not originally developed for continual learning, contain a series of short-to-medium-length videos of the real world collected using wearable glasses.
We note that their dataset has semantic discontinuities, whereas our datasets come from single, continuous video streams.

\section{Conclusion}
\label{sec:conclusion}

This paper is the first to explore lifelong learning of diffusion models from continuous video streams to our knowledge.
We establish that video diffusion models can be trained in a sequential manner from a single autocorrelated and nonstationary video stream using a simple replay-based approach, and that this approach remains effective across data streams, model architectures, parameter sizes, and optimizer choices.
We also introduce five single-video datasets that test how different video stream characteristics affect lifelong learning of video diffusion models.
Training video diffusion models from continuous videos holds promise for never-ending video world model learning for embodied agents \citep{agarwal2025cosmos,alonso2024diffusion,huang2023diffusion,escontrela2024video} in novel environments and compute-efficient foundation model adaptation \citep{smith2024continual}.
Our results suggest that learning generative models from strongly correlated data streams may be less challenging than previously assumed, echoing prior results on discriminative lifelong learning \citep{bornschein2024transformers,carreira2024learning,han2025learning,elsayed2025sequence}.
These investigations lay the groundwork for advancing lifelong video generative modeling by scaling video streams and model sizes, employing advanced diffusion architectures, and building more memory-efficient algorithms for severely memory-constrained settings.

\section*{Acknowledgements}
We acknowledge the support of the Natural Sciences and
Engineering Research Council of Canada (NSERC), the
Canada CIFAR AI Chairs Program, Inverted AI, MITACS,
the Department of Energy through Lawrence Berkeley National Laboratory, and Google. This research was enabled in
part by technical support and computational resources provided by the Digital Research Alliance of Canada Compute
Canada (alliancecan.ca), the Advanced Research Computing at the University of British Columbia (arc.ubc.ca), and
Amazon.

%
%
\bibliographystyle{splncs04}
\bibliography{main}

@String(ICIP  = {IEEE Int. Conf. Image Process.})

@String(ICIP  = {ICIP})

@article{duchi2012ergodic,
  title={Ergodic mirror descent},
  author={Duchi, John C and Agarwal, Alekh and Johansson, Mikael and Jordan, Michael I},
  journal={SIAM Journal on Optimization},
  volume={22},
  number={4},
  pages={1549--1578},
  year={2012},
  publisher={SIAM}
}

@misc{
elsayed2025sequence,
title={Sequence Learning from Continuous Streams of Data},
author={Mohamed Elsayed and A. Rupam Mahmood},
year={2025},
url={https://openreview.net/forum?id=YgvIjdzR4C}
}

@article{harvey2022flexible,
  title={Flexible diffusion modeling of long videos},
  author={Harvey, William and Naderiparizi, Saeid and Masrani, Vaden and Weilbach, Christian and Wood, Frank},
  journal={Advances in Neural Information Processing Systems},
  volume={35},
  pages={27953--27965},
  year={2022}
}

@inproceedings{yang2025egolife,
  title={Egolife: Towards egocentric life assistant},
  author={Yang, Jingkang and Liu, Shuai and Guo, Hongming and Dong, Yuhao and Zhang, Xiamengwei and Zhang, Sicheng and Wang, Pengyun and Zhou, Zitang and Xie, Binzhu and Wang, Ziyue and others},
  booktitle={Proceedings of the Computer Vision and Pattern Recognition Conference},
  pages={28885--28900},
  year={2025}
}

@misc{bornschein2024transformers,
      title={Transformers for Supervised Online Continual Learning}, 
      author={Jorg Bornschein and Yazhe Li and Amal Rannen-Triki},
      year={2024},
      eprint={2403.01554},
      archivePrefix={arXiv},
      primaryClass={cs.LG}
}

@inproceedings{masip2023continual,
      title={Continual Learning of Diffusion Models with Generative Distillation}, 
      author={Sergi Masip and Pau Rodriguez and Tinne Tuytelaars and Gido M. van de Ven},
      booktitle = 	 {Proceedings of The 3rd Conference on Lifelong Learning Agents (CoLLAs)},
      year = 	 {2024},
      publisher =    {PMLR},
}

@article{smith2024continual,
      title={Continual Diffusion: Continual Customization of Text-to-Image Diffusion with C-LoRA}, 
      author={James Seale Smith and Yen-Chang Hsu and Lingyu Zhang and Ting Hua and Zsolt Kira and Yilin Shen and Hongxia Jin},
      year={2024},
      journal={Transactions on Machine Learning Research},
      issn={2835-8856},
      url={https://openreview.net/forum?id=TZdEgwZ6f3},
}

@misc{villa2022vclimb,
      title={vCLIMB: A Novel Video Class Incremental Learning Benchmark}, 
      author={Andrés Villa and Kumail Alhamoud and Juan León Alcázar and Fabian Caba Heilbron and Victor Escorcia and Bernard Ghanem},
      year={2022},
      eprint={2201.09381},
      archivePrefix={arXiv},
      primaryClass={cs.CV}
}

@article{zajac2023exploring,
      title={Exploring Continual Learning of Diffusion Models}, 
      author={Zaj{\k{a}}c, Micha{\l} and Deja, Kamil and Kuzina, Anna and Tomczak, Jakub M and Trzci{\'n}ski, Tomasz and Shkurti, Florian and Mi{\l}o{\'s}, Piotr},
      year={2023},
      journal={arXiv preprint arXiv:2303.15342},
}

@article{videoworldsimulators2024,
  title={Video generation models as world simulators},
  author={Tim Brooks and Bill Peebles and Connor Holmes and Will DePue and Yufei Guo and Li Jing and David Schnurr and Joe Taylor and Troy Luhman and Eric Luhman and Clarence Ng and Ricky Wang and Aditya Ramesh},
  year={2024},
  url={https://openai.com/research/video-generation-models-as-world-simulators},
}

@article{wang2024comprehensive,
  title={A comprehensive survey of continual learning: theory, method and application},
  author={Wang, Liyuan and Zhang, Xingxing and Su, Hang and Zhu, Jun},
  journal={IEEE Transactions on Pattern Analysis and Machine Intelligence},
  year={2024},
  publisher={IEEE},
  volume={46},
  number={8},
  pages={5362-5383},
}

@misc{unterthiner2019accurate,
      title={Towards Accurate Generative Models of Video: A New Metric \& Challenges}, 
      author={Thomas Unterthiner and Sjoerd van Steenkiste and Karol Kurach and Raphael Marinier and Marcin Michalski and Sylvain Gelly},
      year={2019},
      eprint={1812.01717},
      archivePrefix={arXiv},
      primaryClass={cs.CV}
}

@article{vitter1985reservoir,
    author = {Vitter, Jeffrey S.},
    title = {Random sampling with a reservoir},
    year = {1985},
    issue_date = {March 1985},
    publisher = {Association for Computing Machinery},
    address = {New York, NY, USA},
    volume = {11},
    number = {1},
    issn = {0098-3500},
    url = {https://doi.org/10.1145/3147.3165},
    doi = {10.1145/3147.3165},
    journal = {ACM Trans. Math. Softw.},
    month = {mar},
    pages = {37–57},
    numpages = {21}
}

@misc{carreira2024learning,
      title={Learning from One Continuous Video Stream}, 
      author={João Carreira and Michael King and Viorica Pătrăucean and Dilara Gokay and Cătălin Ionescu and Yi Yang and Daniel Zoran and Joseph Heyward and Carl Doersch and Yusuf Aytar and Dima Damen and Andrew Zisserman},
      year={2024},
      eprint={2312.00598},
      archivePrefix={arXiv},
      primaryClass={cs.CV}
}

@misc{rasouli2020deep,
      title={Deep Learning for Vision-based Prediction: A Survey}, 
      author={Amir Rasouli},
      year={2020},
      eprint={2007.00095},
      archivePrefix={arXiv},
      primaryClass={cs.CV}
}

@misc{dprotp2018hours,
 author = {Dprotp},
 year = {2018},
 title = {10 Hours of Windows 3D Maze},
 url = {https://www.youtube.com/watch?v=MHGnSqr9kK8&ab_channel=Dprotp},
 urldate = {2024-05-31}
}

@misc{best2020windows,
 author = {{The Best Classic \& Retro Screensavers}},
 year = {2020},
 title = {Windows 95 Screensaver - 3D Maze - 10 HOURS NO LOOP (4K) NEW},
 url = {https://www.youtube.com/watch?v=Hs5pyyPTzDE&ab_channel=TheBestClassic%26RetroScreensavers},
 urldate = {2024-05-31}
}

@misc{china2023kilometers,
 author = {China Street View},
 year = {2023},
 title = {1674.8 kilometers of long-distance travel across half of China - driving from Chongqing to Shanghai},
 url = {https://www.youtube.com/watch?v=-iSewsdMxkg&ab_channel=%E4%B8%AD%E5%9B%BD%E8%A1%97%E6%99%AFChinaStreetView},
 urldate = {2025-05-10}
}

@article{chaudhry2019tiny,
  title={On tiny episodic memories in continual learning},
  author={Chaudhry, Arslan and Rohrbach, Marcus and Elhoseiny, Mohamed and Ajanthan, Thalaiyasingam and Dokania, Puneet K and Torr, Philip HS and Ranzato, Marc'Aurelio},
  journal={arXiv preprint arXiv:1902.10486},
  year={2019}
}

@article{huang2023diffusion,
  title={Diffusion Reward: Learning Rewards via Conditional Video Diffusion},
  author={Huang, Tao and Jiang, Guangqi and Ze, Yanjie and Xu, Huazhe},
  journal={arXiv preprint arXiv:2312.14134},
  year={2023}
}

@inproceedings{
    lu2024vdt,
    title={{VDT}: General-purpose Video Diffusion Transformers via Mask Modeling},
    author={Haoyu Lu and Guoxing Yang and Nanyi Fei and Yuqi Huo and Zhiwu Lu and Ping Luo and Mingyu Ding},
    booktitle={The Twelfth International Conference on Learning Representations},
    year={2024},
    url={https://openreview.net/forum?id=Un0rgm9f04}
}

@article{escontrela2024video,
  title={Video prediction models as rewards for reinforcement learning},
  author={Escontrela, Alejandro and Adeniji, Ademi and Yan, Wilson and Jain, Ajay and Peng, Xue Bin and Goldberg, Ken and Lee, Youngwoon and Hafner, Danijar and Abbeel, Pieter},
  journal={Advances in Neural Information Processing Systems},
  volume={36},
  year={2024}
}

@inproceedings{rombach2022high,
  title={High-resolution image synthesis with latent diffusion models},
  author={Rombach, Robin and Blattmann, Andreas and Lorenz, Dominik and Esser, Patrick and Ommer, Bj{\"o}rn},
  booktitle={Proceedings of the IEEE/CVF conference on computer vision and pattern recognition},
  pages={10684--10695},
  year={2022}
}

@article{blattmann2023stable,
  title={Stable video diffusion: Scaling latent video diffusion models to large datasets},
  author={Blattmann, Andreas and Dockhorn, Tim and Kulal, Sumith and Mendelevitch, Daniel and Kilian, Maciej and Lorenz, Dominik and Levi, Yam and English, Zion and Voleti, Vikram and Letts, Adam and others},
  journal={arXiv preprint arXiv:2311.15127},
  year={2023}
}

@article{bar2024lumiere,
  title={Lumiere: A space-time diffusion model for video generation},
  author={Bar-Tal, Omer and Chefer, Hila and Tov, Omer and Herrmann, Charles and Paiss, Roni and Zada, Shiran and Ephrat, Ariel and Hur, Junhwa and Li, Yuanzhen and Michaeli, Tomer and others},
  journal={arXiv preprint arXiv:2401.12945},
  year={2024}
}

@inproceedings{blattmann2023align,
  title={Align your latents: High-resolution video synthesis with latent diffusion models},
  author={Blattmann, Andreas and Rombach, Robin and Ling, Huan and Dockhorn, Tim and Kim, Seung Wook and Fidler, Sanja and Kreis, Karsten},
  booktitle={Proceedings of the IEEE/CVF Conference on Computer Vision and Pattern Recognition},
  pages={22563--22575},
  year={2023}
}

@article{yoo2022bayespcn,
  title={Bayespcn: A continually learnable predictive coding associative memory},
  author={Yoo, Jinsoo and Wood, Frank},
  journal={Advances in Neural Information Processing Systems},
  volume={35},
  pages={29903--29914},
  year={2022}
}

@article{zucchet2023online,
  title={Online learning of long-range dependencies},
  author={Zucchet, Nicolas and Meier, Robert and Schug, Simon and Mujika, Asier and Sacramento, Jo{\~a}o},
  journal={Advances in Neural Information Processing Systems},
  volume={36},
  pages={10477--10493},
  year={2023}
}

@inproceedings{
nguyen2018variational,
title={Variational Continual Learning},
author={Cuong V. Nguyen and Yingzhen Li and Thang D. Bui and Richard E. Turner},
booktitle={International Conference on Learning Representations},
year={2018},
url={https://openreview.net/forum?id=BkQqq0gRb},
}

@article{ramapuram2020lifelong,
  title={Lifelong generative modeling},
  author={Ramapuram, Jason and Gregorova, Magda and Kalousis, Alexandros},
  journal={Neurocomputing},
  volume={404},
  pages={381--400},
  year={2020},
  publisher={Elsevier}
}

@inproceedings{chen2022continual,
  title={Continual predictive learning from videos},
  author={Chen, Geng and Zhang, Wendong and Lu, Han and Gao, Siyu and Wang, Yunbo and Long, Mingsheng and Yang, Xiaokang},
  booktitle={Proceedings of the IEEE/CVF Conference on Computer Vision and Pattern Recognition},
  pages={10728--10737},
  year={2022}
}

@article{shin2017continual,
  title={Continual learning with deep generative replay},
  author={Shin, Hanul and Lee, Jung Kwon and Kim, Jaehong and Kim, Jiwon},
  journal={Advances in neural information processing systems},
  volume={30},
  year={2017}
}

@article{hoppe2022diffusion,
title={Diffusion Models for Video Prediction and Infilling},
author={Tobias H{\"o}ppe and Arash Mehrjou and Stefan Bauer and Didrik Nielsen and Andrea Dittadi},
journal={Transactions on Machine Learning Research},
issn={2835-8856},
year={2022},
url={https://openreview.net/forum?id=lf0lr4AYM6},
note={}
}

@article{green2024semantically,
  title={Semantically Consistent Video Inpainting with Conditional Diffusion Models},
  author={Green, Dylan and Harvey, William and Naderiparizi, Saeid and Niedoba, Matthew and Liu, Yunpeng and Liang, Xiaoxuan and Lavington, Jonathan and Zhang, Ke and Lioutas, Vasileios and Dabiri, Setareh and others},
  journal={arXiv preprint arXiv:2405.00251},
  year={2024}
}

@article{ho2022video,
  title={Video diffusion models},
  author={Ho, Jonathan and Salimans, Tim and Gritsenko, Alexey and Chan, William and Norouzi, Mohammad and Fleet, David J},
  journal={Advances in Neural Information Processing Systems},
  volume={35},
  pages={8633--8646},
  year={2022}
}

@article{voleti2022mcvd,
  title={Mcvd-masked conditional video diffusion for prediction, generation, and interpolation},
  author={Voleti, Vikram and Jolicoeur-Martineau, Alexia and Pal, Chris},
  journal={Advances in neural information processing systems},
  volume={35},
  pages={23371--23385},
  year={2022}
}

@inproceedings{campo2020continual,
  title={Continual learning of predictive models in video sequences via variational autoencoders},
  author={Campo, Damian and Slavic, Giulia and Baydoun, Mohamad and Marcenaro, Lucio and Regazzoni, Carlo},
  booktitle={2020 IEEE International Conference on Image Processing (ICIP)},
  pages={753--757},
  year={2020},
  organization={IEEE}
}

@article{tang2024vilco,
  title={ViLCo-Bench: VIdeo Language COntinual learning Benchmark},
  author={Tang, Tianqi and Deldari, Shohreh and Xue, Hao and De Melo, Celso and Salim, Flora D},
  journal={arXiv preprint arXiv:2406.13123},
  year={2024}
}

@INPROCEEDINGS{singh2016krishnacam,
  author={Singh, Krishna Kumar and Fatahalian, Kayvon and Efros, Alexei A.},
  booktitle={2016 IEEE Winter Conference on Applications of Computer Vision (WACV)}, 
  title={KrishnaCam: Using a longitudinal, single-person, egocentric dataset for scene understanding tasks}, 
  year={2016},
  volume={},
  number={},
  pages={1-9},
  doi={10.1109/WACV.2016.7477717}
}

@inproceedings{yoo2024layerwise,
  title={Layerwise Proximal Replay: A Proximal Point Method for Online Continual Learning},
  author={Jinsoo Yoo and Yunpeng Liu and Frank Wood and Geoff Pleiss},
  booktitle={Forty-first International Conference on Machine Learning},
  year={2024},
  url={https://openreview.net/forum?id=Lg8nw3ltvX}
}

@InProceedings{hess2023two,
  title = 	 {Two Complementary Perspectives to Continual Learning: Ask Not Only What to Optimize, But Also How},
  author =       {Hess, Timm and Tuytelaars, Tinne and van de Ven, Gido M},
  booktitle = 	 {Proceedings of the 1st ContinualAI Unconference},
  pages = 	 {37--61},
  year = 	 {2023},
  volume = 	 {249},
  series = 	 {Proceedings of Machine Learning Research},
  publisher =    {PMLR},
  url = 	 {https://proceedings.mlr.press/v249/hess24a.html},
}

@article{hemati2024continual,
  title={Continual Learning in the Presence of Repetition},
  author={Hemati, Hamed and Pellegrini, Lorenzo and Duan, Xiaotian and Zhao, Zixuan and Xia, Fangfang and Masana, Marc and Tscheschner, Benedikt and Veas, Eduardo and Zheng, Yuxiang and Zhao, Shiji and others},
  journal={Neural Networks},
  volume={183},
  pages={106920},
  year={2025}
}

@article{dohare2024loss,
  title={Loss of plasticity in deep continual learning},
  author={Dohare, Shibhansh and Hernandez-Garcia, J Fernando and Lan, Qingfeng and Rahman, Parash and Mahmood, A Rupam and Sutton, Richard S},
  journal={Nature},
  volume={632},
  number={8026},
  pages={768--774},
  year={2024},
  publisher={Nature Publishing Group UK London}
}

@article{van2024continual,
  title={Continual Learning and Catastrophic Forgetting},
  author={van de Ven, Gido M and Soures, Nicholas and Kudithipudi, Dhireesha},
  journal={arXiv preprint arXiv:2403.05175},
  year={2024}
}

@InProceedings{zenke2017continual,
  title = 	 {Continual Learning Through Synaptic Intelligence},
  author =       {Friedemann Zenke and Ben Poole and Surya Ganguli},
  booktitle = 	 {Proceedings of the 34th International Conference on Machine Learning},
  pages = 	 {3987--3995},
  year = 	 {2017},
  volume = 	 {70},
  series = 	 {Proceedings of Machine Learning Research},
  publisher =    {PMLR},
  url = 	 {https://proceedings.mlr.press/v70/zenke17a.html},
}

@article{kirkpatrick2017overcoming,
  title={Overcoming catastrophic forgetting in neural networks},
  author={Kirkpatrick, James and Pascanu, Razvan and Rabinowitz, Neil and Veness, Joel and Desjardins, Guillaume and Rusu, Andrei A and Milan, Kieran and Quan, John and Ramalho, Tiago and Grabska-Barwinska, Agnieszka and others},
  journal={Proceedings of the National Academy of Sciences},
  volume={114},
  number={13},
  pages={3521--3526},
  year={2017},
  publisher={National Acad Sciences}
}

@article{li2017learning,
  title={Learning without forgetting},
  author={Li, Zhizhong and Hoiem, Derek},
  journal={IEEE Transactions on Pattern Analysis and Machine Intelligence},
  volume={40},
  number={12},
  pages={2935--2947},
  year={2017},
  publisher={IEEE}
}

@article{rolnick2019experience,
  title={Experience replay for continual learning},
  author={Rolnick, David and Ahuja, Arun and Schwarz, Jonathan and Lillicrap, Timothy and Wayne, Gregory},
  journal={Advances in Neural Information Processing Systems},
  volume={32},
  year={2019}
}

@article{robins1995catastrophic,
  title={Catastrophic forgetting, rehearsal and pseudorehearsal},
  author={Robins, Anthony},
  journal={Connection Science},
  volume={7},
  number={2},
  pages={123--146},
  year={1995},
  publisher={Taylor \& Francis}
}

@ARTICLE{delange2022continual,
  author={De Lange, Matthias and Aljundi, Rahaf and Masana, Marc and Parisot, Sarah and Jia, Xu and Leonardis, Aleš and Slabaugh, Gregory and Tuytelaars, Tinne},
  journal={IEEE Transactions on Pattern Analysis and Machine Intelligence}, 
  title={A Continual Learning Survey: Defying Forgetting in Classification Tasks},
  year={2022},
  volume={44},
  number={7},
  pages={3366-3385},
}

@article{zeng2019continual,
  title={Continual learning of context-dependent processing in neural networks},
  author={Zeng, Guanxiong and Chen, Yang and Cui, Bo and Yu, Shan},
  journal={Nature Machine Intelligence},
  volume={1},
  number={8},
  pages={364--372},
  year={2019},
  publisher={Nature Publishing Group UK London}
}

@inproceedings{serra2018overcoming,
  title={Overcoming catastrophic forgetting with hard attention to the task},
  author={Serra, Joan and Suris, Didac and Miron, Marius and Karatzoglou, Alexandros},
  booktitle={International conference on machine learning},
  pages={4548--4557},
  year={2018},
  organization={PMLR}
}

@article{rusu2016progressive,
  title={Progressive neural networks},
  author={Rusu, Andrei A and Rabinowitz, Neil C and Desjardins, Guillaume and Soyer, Hubert and Kirkpatrick, James and Kavukcuoglu, Koray and Pascanu, Razvan and Hadsell, Raia},
  journal={arXiv preprint arXiv:1606.04671},
  year={2016}
}

@article{liu2024locality,
  title={Locality Sensitive Sparse Encoding for Learning World Models Online},
  author={Liu, Zichen and Du, Chao and Lee, Wee Sun and Lin, Min},
  journal={arXiv preprint arXiv:2401.13034},
  year={2024}
}

@inproceedings{zhai2019lifelong,
  title={Lifelong gan: Continual learning for conditional image generation},
  author={Zhai, Mengyao and Chen, Lei and Tung, Frederick and He, Jiawei and Nawhal, Megha and Mori, Greg},
  booktitle={Proceedings of the IEEE/CVF International Conference on Computer Vision},
  pages={2759--2768},
  year={2019}
}

@inproceedings{lesort2019generative,
  title={Generative models from the perspective of continual learning},
  author={Lesort, Timoth{\'e}e and Caselles-Dupr{\'e}, Hugo and Garcia-Ortiz, Michael and Stoian, Andrei and Filliat, David},
  booktitle={2019 International Joint Conference on Neural Networks (IJCNN)},
  pages={1--8},
  year={2019},
  organization={IEEE}
}

@article{egorov2021boovae,
  title={BooVAE: Boosting approach for continual learning of VAE},
  author={Egorov, Evgenii and Kuzina, Anna and Burnaev, Evgeny},
  journal={Advances in Neural Information Processing Systems},
  volume={34},
  pages={17889--17901},
  year={2021}
}

@article{chen2020mitigating,
  title={Mitigating forgetting in online continual learning via instance-aware parameterization},
  author={Chen, Hung-Jen and Cheng, An-Chieh and Juan, Da-Cheng and Wei, Wei and Sun, Min},
  journal={Advances in Neural Information Processing Systems},
  volume={33},
  pages={17466--17477},
  year={2020}
}

@article{aljundi2019gradient,
  title={Gradient based sample selection for online continual learning},
  author={Aljundi, Rahaf and Lin, Min and Goujaud, Baptiste and Bengio, Yoshua},
  journal={Advances in Neural Information Processing Systems},
  volume={32},
  year={2019}
}

@article{verwimp2023clad,
  title={{CLAD}: A realistic continual learning benchmark for autonomous driving},
  author={Verwimp, Eli and Yang, Kuo and Parisot, Sarah and Hong, Lanqing and McDonagh, Steven and P{\'e}rez-Pellitero, Eduardo and De Lange, Matthias and Tuytelaars, Tinne},
  journal={Neural Networks},
  volume={161},
  pages={659--669},
  year={2023},
  publisher={Elsevier}
}

@article{
verwimp2024continual,
title={Continual Learning: Applications and the Road Forward},
author={Eli Verwimp and Rahaf Aljundi and Shai Ben-David and Matthias Bethge and Andrea Cossu and Alexander Gepperth and Tyler L. Hayes and Eyke H{\"u}llermeier and Christopher Kanan and Dhireesha Kudithipudi and Christoph H. Lampert and Martin Mundt and Razvan Pascanu and Adrian Popescu and Andreas S. Tolias and Joost van de Weijer and Bing Liu and Vincenzo Lomonaco and Tinne Tuytelaars and Gido M van de Ven},
journal={Transactions on Machine Learning Research},
issn={2835-8856},
year={2024},
url={https://openreview.net/forum?id=axBIMcGZn9},
note={}
}

@article{bornschein2023nevis,
  title={Nevis' 22: A Stream of 100 Tasks Sampled from 30 Years of Computer Vision Research},
  author={Bornschein, Jorg and Galashov, Alexandre and Hemsley, Ross and Rannen-Triki, Amal and Chen, Yutian and Chaudhry, Arslan and He, Xu Owen and Douillard, Arthur and Caccia, Massimo and Feng, Qixuan and others},
  journal={Journal of Machine Learning Research},
  volume={24},
  number={308},
  pages={1--77},
  year={2023}
}

@inproceedings{bang2022online,
  title={Online continual learning on a contaminated data stream with blurry task boundaries},
  author={Bang, Jihwan and Koh, Hyunseo and Park, Seulki and Song, Hwanjun and Ha, Jung-Woo and Choi, Jonghyun},
  booktitle={Proceedings of the IEEE/CVF Conference on Computer Vision and Pattern Recognition},
  pages={9275--9284},
  year={2022}
}

@inproceedings{moon2023online,
  title={Online class incremental learning on stochastic blurry task boundary via mask and visual prompt tuning},
  author={Moon, Jun-Yeong and Park, Keon-Hee and Kim, Jung Uk and Park, Gyeong-Moon},
  booktitle={Proceedings of the IEEE/CVF International Conference on Computer Vision},
  pages={11731--11741},
  year={2023}
}

@article{buzzega2020dark,
  title={Dark experience for general continual learning: a strong, simple baseline},
  author={Buzzega, Pietro and Boschini, Matteo and Porrello, Angelo and Abati, Davide and Calderara, Simone},
  journal={Advances in Neural Information Processing Systems},
  volume={33},
  pages={15920--15930},
  year={2020}
}

@inproceedings{arani2022learning,
  title={Learning Fast, Learning Slow: A General Continual Learning Method based on Complementary Learning System},
  author={Elahe Arani and Fahad Sarfraz and Bahram Zonooz},
  booktitle={International Conference on Learning Representations},
  year={2022},
  url={https://openreview.net/forum?id=uxxFrDwrE7Y}
}

@article{touvron2023llama,
  title={Llama: Open and efficient foundation language models},
  author={Touvron, Hugo and Lavril, Thibaut and Izacard, Gautier and Martinet, Xavier and Lachaux, Marie-Anne and Lacroix, Timoth{\'e}e and Rozi{\`e}re, Baptiste and Goyal, Naman and Hambro, Eric and Azhar, Faisal and others},
  journal={arXiv preprint arXiv:2302.13971},
  year={2023}
}

@article{mukherjee2023orca,
  title={Orca: Progressive learning from complex explanation traces of gpt-4},
  author={Mukherjee, Subhabrata and Mitra, Arindam and Jawahar, Ganesh and Agarwal, Sahaj and Palangi, Hamid and Awadallah, Ahmed},
  journal={arXiv preprint arXiv:2306.02707},
  year={2023}
}

@inproceedings{
hafner2020dream,
title={Dream to Control: Learning Behaviors by Latent Imagination},
author={Danijar Hafner and Timothy Lillicrap and Jimmy Ba and Mohammad Norouzi},
booktitle={International Conference on Learning Representations},
year={2020},
url={https://openreview.net/forum?id=S1lOTC4tDS}
}

@inproceedings{
hafner2021mastering,
title={Mastering Atari with Discrete World Models},
author={Danijar Hafner and Timothy P Lillicrap and Mohammad Norouzi and Jimmy Ba},
booktitle={International Conference on Learning Representations},
year={2021},
url={https://openreview.net/forum?id=0oabwyZbOu}
}

@article{
godichon-baggioni2023learning,
title={Learning from time-dependent streaming data with online stochastic algorithms},
author={Antoine Godichon-Baggioni and Nicklas Werge and Olivier Wintenberger},
journal={Transactions on Machine Learning Research},
issn={2835-8856},
year={2023},
url={https://openreview.net/forum?id=kdfiEu1ul6},
note={}
}

@InProceedings{pmlr-v161-beronov21a,
  title = 	 {Sequential core-set Monte Carlo},
  author =       {Beronov, Boyan and Weilbach, Christian and Wood, Frank and Campbell, Trevor},
  booktitle = 	 {Proceedings of the Thirty-Seventh Conference on Uncertainty in Artificial Intelligence},
  pages = 	 {2165--2175},
  year = 	 {2021},
  editor = 	 {de Campos, Cassio and Maathuis, Marloes H.},
  volume = 	 {161},
  series = 	 {Proceedings of Machine Learning Research},
  month = 	 {27--30 Jul},
  publisher =    {PMLR},
  url = 	 {https://proceedings.mlr.press/v161/beronov21a.html},
}

@article{naesseth2019elements,
  title={Elements of sequential monte carlo},
  author={Naesseth, Christian A and Lindsten, Fredrik and Sch{\"o}n, Thomas B and others},
  journal={Foundations and Trends{\textregistered} in Machine Learning},
  volume={12},
  number={3},
  pages={307--392},
  year={2019},
  publisher={Now Publishers, Inc.}
}

@article{broderick2013streaming,
  title={Streaming variational bayes},
  author={Broderick, Tamara and Boyd, Nicholas and Wibisono, Andre and Wilson, Ashia C and Jordan, Michael I},
  journal={Advances in neural information processing systems},
  volume={26},
  year={2013}
}

@INPROCEEDINGS{bartlett2011deplump,
  author={Bartlett, Nicholas and Wood, Frank},
  booktitle={2011 Data Compression Conference}, 
  title={Deplump for Streaming Data}, 
  year={2011},
  volume={},
  number={},
  pages={363-372},
  doi={10.1109/DCC.2011.43}}

@article{Lillicrap2020,
  title = {Backpropagation and the brain},
  volume = {21},
  ISSN = {1471-0048},
  url = {http://dx.doi.org/10.1038/s41583-020-0277-3},
  DOI = {10.1038/s41583-020-0277-3},
  number = {6},
  journal = {Nature Reviews Neuroscience},
  publisher = {Springer Science and Business Media LLC},
  author = {Lillicrap,  Timothy P. and Santoro,  Adam and Marris,  Luke and Akerman,  Colin J. and Hinton,  Geoffrey},
  year = {2020},
  month = apr,
  pages = {335–346}
}

@article{kowshik2021streaming,
  title={Streaming linear system identification with reverse experience replay},
  author={Kowshik, Suhas and Nagaraj, Dheeraj and Jain, Prateek and Netrapalli, Praneeth},
  journal={Advances in Neural Information Processing Systems},
  volume={34},
  pages={30140--30152},
  year={2021}
}

@article{xu2023parameter,
  title={Parameter-efficient fine-tuning methods for pretrained language models: A critical review and assessment},
  author={Xu, Lingling and Xie, Haoran and Qin, Si-Zhao Joe and Tao, Xiaohui and Wang, Fu Lee},
  journal={arXiv preprint arXiv:2312.12148},
  year={2023}
}

@article{ash2020warm,
  title={On warm-starting neural network training},
  author={Ash, Jordan and Adams, Ryan P},
  journal={Advances in neural information processing systems},
  volume={33},
  pages={3884--3894},
  year={2020}
}

@article{ren2021survey,
  title={A survey of deep active learning},
  author={Ren, Pengzhen and Xiao, Yun and Chang, Xiaojun and Huang, Po-Yao and Li, Zhihui and Gupta, Brij B and Chen, Xiaojiang and Wang, Xin},
  journal={ACM computing surveys (CSUR)},
  volume={54},
  number={9},
  pages={1--40},
  year={2021},
  publisher={ACM New York, NY}
}

@article{ho2020denoising,
  title={Denoising diffusion probabilistic models},
  author={Ho, Jonathan and Jain, Ajay and Abbeel, Pieter},
  journal={Advances in neural information processing systems},
  volume={33},
  pages={6840--6851},
  year={2020}
}

@article{karras2022elucidating,
  title={Elucidating the design space of diffusion-based generative models},
  author={Karras, Tero and Aittala, Miika and Aila, Timo and Laine, Samuli},
  journal={Advances in neural information processing systems},
  volume={35},
  pages={26565--26577},
  year={2022}
}

@article{johansson2010randomized,
author = {Johansson, Bj\"{o}rn and Rabi, Maben and Johansson, Mikael},
title = {A Randomized Incremental Subgradient Method for Distributed Optimization in Networked Systems},
journal = {SIAM Journal on Optimization},
volume = {20},
number = {3},
pages = {1157-1170},
year = {2010},
doi = {10.1137/08073038X},
URL = { 
        https://doi.org/10.1137/08073038X
},
eprint = { 
        https://doi.org/10.1137/08073038X
}
,
}

@article{ram2009incremental,
  title={Incremental stochastic subgradient algorithms for convex optimization},
  author={Ram, S Sundhar and Nedi{\'c}, A and Veeravalli, Venugopal V},
  journal={SIAM Journal on Optimization},
  volume={20},
  number={2},
  pages={691--717},
  year={2009},
  publisher={SIAM}
}

@inproceedings{chang2022distributed,
  title={Distributed online system identification for lti systems using reverse experience replay},
  author={Chang, Ting-Jui and Shahrampour, Shahin},
  booktitle={2022 IEEE 61st Conference on Decision and Control (CDC)},
  pages={6672--6677},
  year={2022},
  organization={IEEE}
}

@article{he2025plaicraft,
  title={PLAICraft: Large-Scale Time-Aligned Vision-Speech-Action Dataset for Embodied AI},
  author={He, Yingchen and Weilbach, Christian D and Wojciechowska, Martyna E and Zhang, Yuxuan and Wood, Frank},
  journal={arXiv preprint arXiv:2505.12707},
  year={2025}
}

@article{alonso2024diffusion,
  title={Diffusion for World Modeling: Visual Details Matter in Atari},
  author={Alonso, Eloi and Jelley, Adam and Micheli, Vincent and Kanervisto, Anssi and Storkey, Amos and Pearce, Tim and Fleuret, Fran{\c{c}}ois},
  journal={arXiv preprint arXiv:2405.12399},
  year={2024}
}

@inproceedings{
luo2025beyond,
title={Beyond {FVD}: An Enhanced Evaluation Metrics for Video Generation Distribution Quality},
author={Ge Ya Luo and Gian Mario Favero and ZhiHao Luo and Alexia Jolicoeur-Martineau and Christopher Pal},
booktitle={The Thirteenth International Conference on Learning Representations},
year={2025},
url={https://openreview.net/forum?id=cC3LxGZasH}
}

@article{han2025learning,
  title={Learning from Streaming Video with Orthogonal Gradients},
  author={Han, Tengda and Gokay, Dilara and Heyward, Joseph and Zhang, Chuhan and Zoran, Daniel and P{\u{a}}tr{\u{a}}ucean, Viorica and Carreira, Jo{\~a}o and Damen, Dima and Zisserman, Andrew},
  journal={arXiv preprint arXiv:2504.01961},
  year={2025}
}

@article{agarwal2025cosmos,
  title={Cosmos world foundation model platform for physical ai},
  author={Agarwal, Niket and Ali, Arslan and Bala, Maciej and Balaji, Yogesh and Barker, Erik and Cai, Tiffany and Chattopadhyay, Prithvijit and Chen, Yongxin and Cui, Yin and Ding, Yifan and others},
  journal={arXiv preprint arXiv:2501.03575},
  year={2025}
}

@article{liu2025can,
  title={How can Diffusion Models Evolve into Continual Generators?},
  author={Liu, Jingren and Ji, Zhong and Chen, Xiangyu},
  journal={arXiv preprint arXiv:2505.11936},
  year={2025}
}

@inproceedings{
loshchilov2018decoupled,
title={Decoupled Weight Decay Regularization},
author={Ilya Loshchilov and Frank Hutter},
booktitle={International Conference on Learning Representations},
year={2019},
url={https://openreview.net/forum?id=Bkg6RiCqY7},
}

\appendix
\crefalias{section}{appendix}

\newpage
\appendix

\section{The ColorKL Metric}
\label{appendix:metrics}

The ColorKL metric measures how faithful a model-generated video's color transition statistics are to the ground truth color transition statistics at the dataset level.
It is defined as
\begin{gather}
    \frac{1}{|\mathcal{C}|} \sum_{c_{old} \in \mathcal{C}} D_\mathrm{KL} \left(p_*(c_{new}|c_{old}) \;\|\ p_{model}(c_{new}|c_{old})\right)
\end{gather}
where $D_\mathrm{KL}$ is the KL divergence, $\mathcal{C} = \{red, yellow, green\}$ is the set of possible ball colors, $p_*(c_{new}|c_{old})$ is the ground truth probability for a ball to switch colors from $c_{old}$ to $c_{new}$, and $p_{model}$ is the empirical probability of color transitions from $c_{old}$ to $c_{new}$ in the model-generated video dataset.
We note that if the model's context window is long enough to always capture two past ball bounces for all balls, the ball colors can be deterministically predicted.
If the model's context is not long enough, as is the case in our experiments, the ground truth transition probabilities $p_*$ state that the balls always transition from yellow to red and green to red but have a 50/50 chance of transitioning from red to yellow or green.

\section{Additional Experiment Details}
\label{appendix:expdetails}


The minimum average displacement error (minADE) metric is computed by selecting the trajectory with the lowest average displacement error from 3 sampled trajectories for each evaluation video subsequence.
The ColorKL metric is computed by tallying the transition statistics from 3 sampled trajectories for each evaluation video subsequence and computing the KL divergence of this empirical distribution with the ground truth transition statistics.
The loss metric is calculated by uniformly sampling 10 noise levels for all evaluation set samples per model checkpoint, computing the diffusion MSE loss, and averaging the results.

As computing all metrics on the entirety of the video streams is prohibitively expensive, we select 1,000 video subsequences from the train and test streams and calculate the metrics on those video frames for all reported metrics.
All training and sampling seeds compute the metrics on the same set of video subsequences.
For datasets without significant changes in video frame details throughout
the video stream (\textit{Lifelong Bouncing Balls (O)}), we select the first 1,000 video subsequences from the evaluation video stream.
For datasets with significant changes in video frame details throughout the video stream (\textit{Lifelong Bouncing Balls (C), Lifelong Drive, Lifelong PLAICraft}), we evenly select 1,000 video subsequences across the entire evaluation video stream with equal spacing.
For \textit{Lifelong 3D Maze}, the train stream evaluation is done across 1,000 video subsequences evenly selected from the stream and the test stream evaluation is done across the first 1,000 video subsequences from the stream due to the subtle train stream perceptual changes detailed in \Cref{appendix:curation}.

All models are optimized with AdamW with learning rate 1e-4.
The optimizers for U-Net models have weight decay of 1e-5, whereas the optimizers for Transformer models have weight decay of 0 for \textit{Lifelong Bouncing Balls} and 1e-6 for all other experiments.
All U-Net models have the \textit{num\_res\_blocks} hyperparameter set to 1.
The \textit{num\_channels} hyperparameter is set to 64, 192, 128, 128, 128 for \textit{Lifelong Bouncing Balls (O), Lifelong Bouncing Balls (C), Lifelong 3D Maze, Lifelong Drive,} and \textit{Lifelong PLAICraft} experiments respectively.
All VDT models have the \textit{num\_layers} hyperparameter set to 12.
The \textit{hidden\_layer\_size} and \textit{patch\_size} hyperparameters are set to 384, 384, 640, 1024, 1024 and  2, 2, 4, 4, 8 for \textit{Lifelong Bouncing Balls (O), Lifelong Bouncing Balls (C), Lifelong 3D Maze, Lifelong Drive,} and \textit{Lifelong PLAICraft} experiments respectively.
We applied gradient clipping with a threshold of 1 to lifelong learned VDT models for Lifelong PLAICraft and Lifelong Drive experiments to mitigate infrequent but sporadic gradient norm spikes.

The U-Net and VDT experiments for \textit{Lifelong Bouncing Balls} took 3 days to complete on a single GeForce RTX 2080 GPU.
The U-Net and VDT experiments for \textit{Lifelong 3D Maze} took 7 days to complete on two Tesla V100 GPUs.
The U-Net experiments for \textit{Lifelong Drive} and \textit{Lifelong PLAICraft} took 4 days to complete on 4 A5000 GPUs, and the VDT experiments for \textit{Lifelong Drive} and \textit{Lifelong PLAICraft} took 6 days to complete on 4 L40 GPUs.
There was no substantial runtime difference between Offline Learning and Lifelong Learning since both methods employ the same batch sizes and the same number of gradient steps.

\section{Additional Dataset Curation Details}
\label{appendix:curation}

\paragraph{Lifelong Bouncing Balls} videos were created by randomly sampling the two balls' initial positions and velocities and deterministically updating them to satisfy the conservation of momentum.

\paragraph{Lifelong 3D Maze} was created by concatenating two 10-hour-long Lifelong 3D Maze YouTube videos \citep{dprotp2018hours,best2020windows} at the point where a maze from the first video is solved and the maze from the second video begins, as the maze screensaver teleports the agent to a newly generated maze on the completion of the previous maze. This concatenation ensures that there is no sudden switch in the environment dynamics (a slight perceptual switch exists as shown in \Cref{fig:maze-frame-comparison}).

\begin{figure}[ht!]
    \centering
    \begin{subfigure}[c]{0.4\linewidth}
        \centering
        \includegraphics[width=0.4\linewidth]{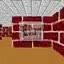}
        \caption{Example frame from video 1.}
    \end{subfigure}
    \begin{subfigure}[c]{0.4\linewidth}
        \centering
        \includegraphics[width=0.4\linewidth]{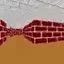}
        \caption{Example frame from video 1.}
    \end{subfigure}
    \\
    \begin{subfigure}[c]{0.4\linewidth}
        \centering
        \includegraphics[width=0.4\linewidth]{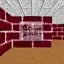}
        \caption{Example frame from video 2.}
    \end{subfigure}
    \begin{subfigure}[c]{0.4\linewidth}
        \centering
        \includegraphics[width=0.4\linewidth]{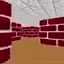}
        \caption{Example frame from video 2.}
    \end{subfigure}
    \caption{Example video frames from first and second 10-hour-long YouTube videos used to construct \textit{Lifelong 3D Maze}. While the frames are nearly identical, there are very subtle perceptual differences arising from how the uploaders recorded the two videos.}
    \label{fig:maze-frame-comparison}
\end{figure}

\paragraph{Lifelong Drive} was constructed by first downloading a YouTube video that features a vehicle driving on a series of highways from Chongqing to Shanghai \citep{china2023kilometers}, then cropping and downsampling the video such that its frame dimensions became 512x512.
The drive is nearly continuous, with subtle and rare discontinuities occurring when the driver takes breaks in between very long drives.
In these scenarios, the video either resumes at the same spot where the vehicle was parked or at a nearby highway location.

\paragraph{Lifelong PLAICraft} was constructed by concatenating multiple consecutive gameplay session recordings from the same player of the PLAICraft server \citep{he2025plaicraft}. All game settings, such as viewing distance, shader configurations, and recording parameters (e.g., resolution and aspect ratio), were consistent across sessions. The dataset includes recordings of two players, Alex and Kyrie, who played in the same survival world alongside hundreds of other players. In their initial gameplay sessions, their starting locations were randomly assigned but could later be altered through travel or teleportation. As in a typical Minecraft survival world, the environment had no boundaries; landscapes were procedurally generated and remained unchanged once created. Player states, including inventories and spawn locations, were preserved between consecutive sessions.

Players’ locations and states were generally consistent across most consecutive gameplay sessions.
In other words, if a player exited the game at a particular location and state in one session, they would typically resume from the same point in the next session. However, some aspects, such as the player's viewing angle or environment, may be slightly different from reasons such as other players having built structures around the location the player logged off from or other technical reasons related to gameplay recording.

Kyrie’s first gameplay session began later than Alex's, and there was some overlap in the locations they visited. Alex generally explored more areas and contributed extensively to building structures, whereas Kyrie primarily played in a village that Alex and other players had previously developed.

\section{Additional Qualitative Results}
\label{appendix:additional_samples}

\Cref{fig:all-comparisons-appendix} illustrates the similarity between models trained from Offline Learning and Lifelong Learning.

\begin{figure}[H]
    \centering
    \begin{minipage}[t]{0.49\linewidth}
        \begin{subfigure}{\linewidth}
        \centering
        \includegraphics[width=6.1cm, height=3.05cm, keepaspectratio=false]{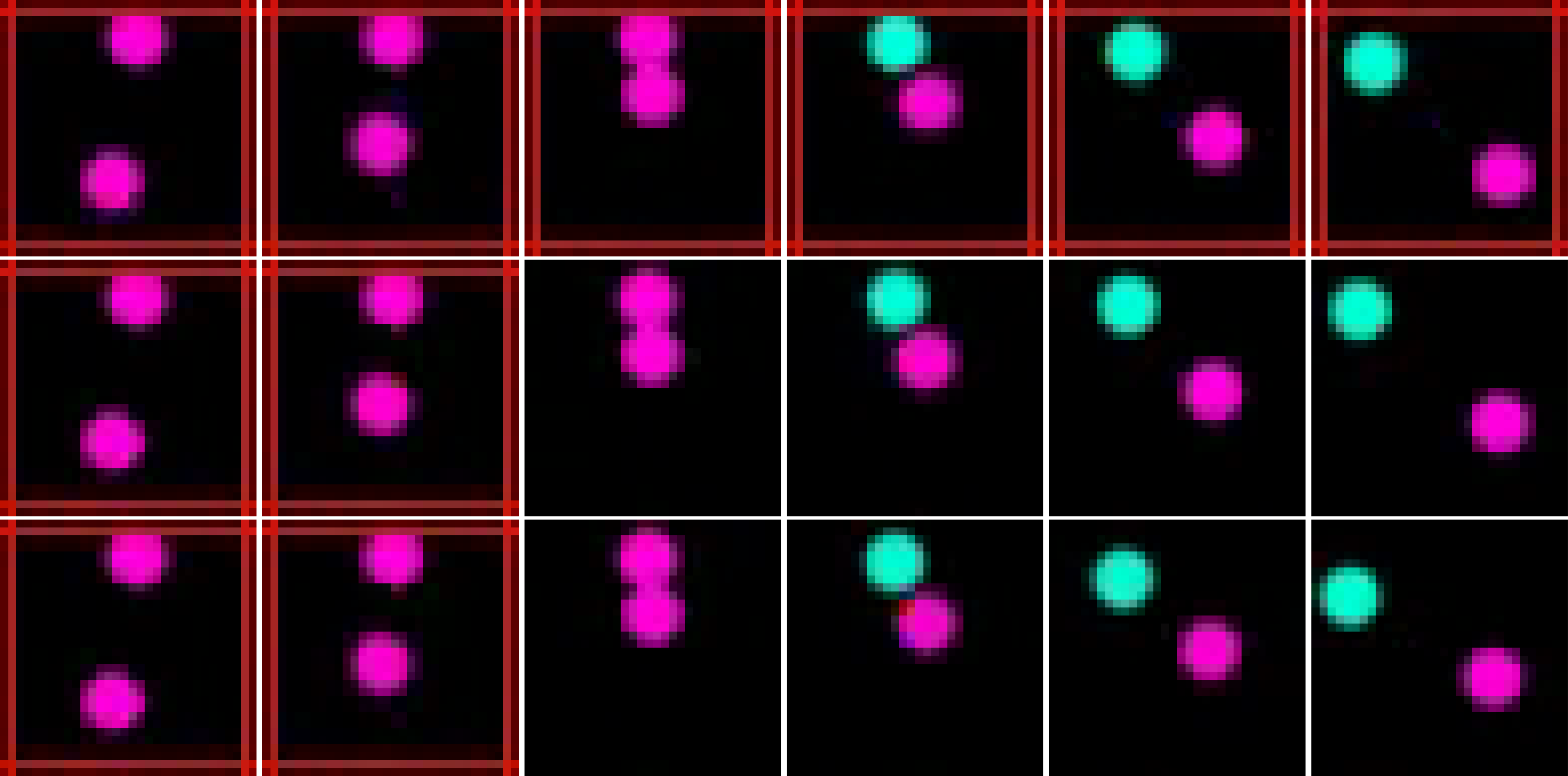}
        \caption{\textit{Lifelong Bouncing Balls (C)} videos.}
        \label{fig:ball-comparison-nstn}
        \end{subfigure}
    \end{minipage}
    \hfill
    \begin{minipage}[t]{0.49\linewidth}
        \begin{subfigure}{\linewidth}
        \centering
        \includegraphics[width=6.1cm, height=3.05cm, keepaspectratio=false]{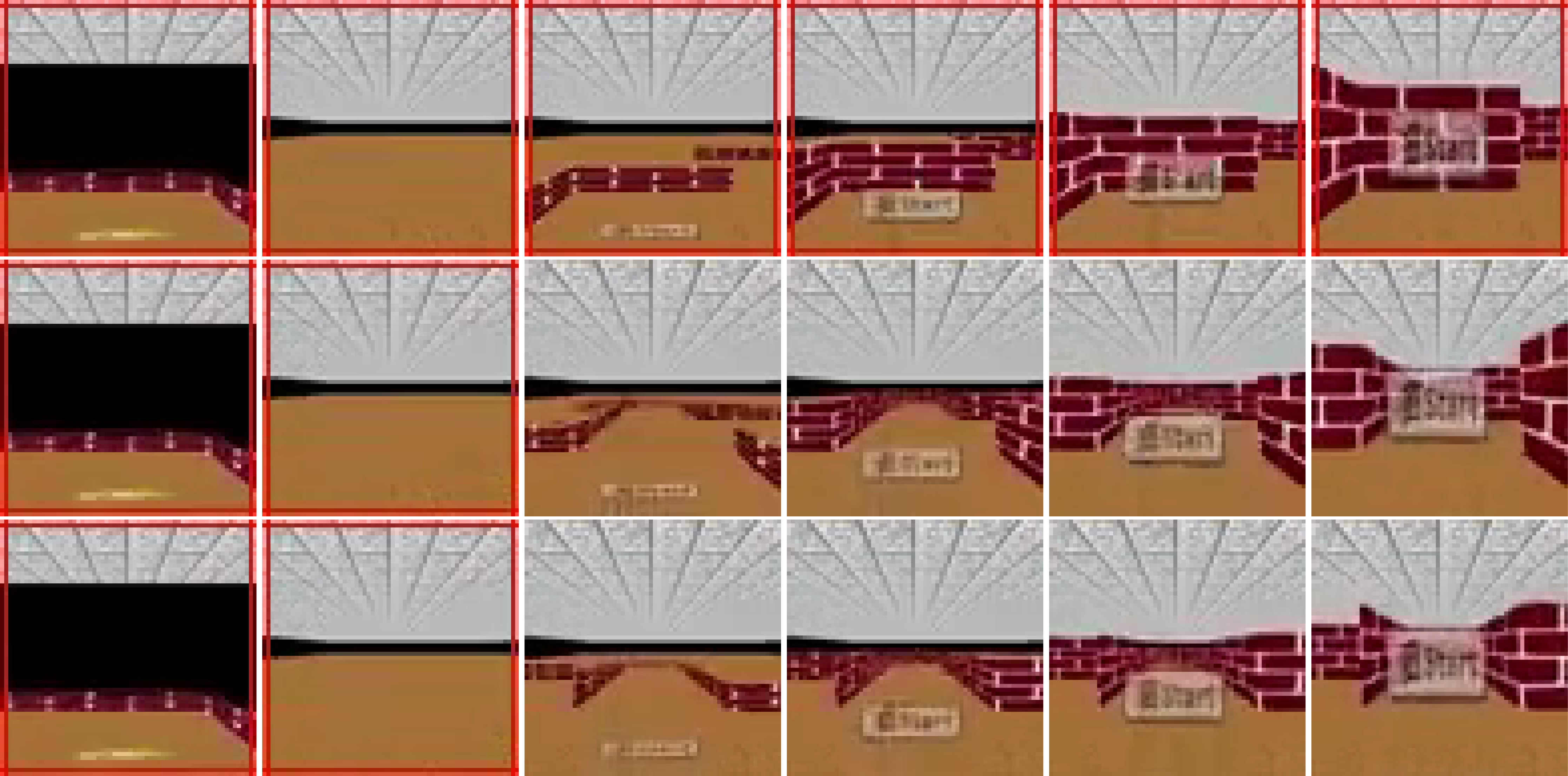}
        \caption{\textit{Lifelong 3D Maze} videos.}
        \label{fig:maze-comparison-rise}
        \end{subfigure}
    \end{minipage}

    \begin{minipage}[t]{0.49\linewidth}
        \begin{subfigure}{\linewidth}
        \centering
        \includegraphics[width=6.1cm, height=3.05cm, keepaspectratio=false]{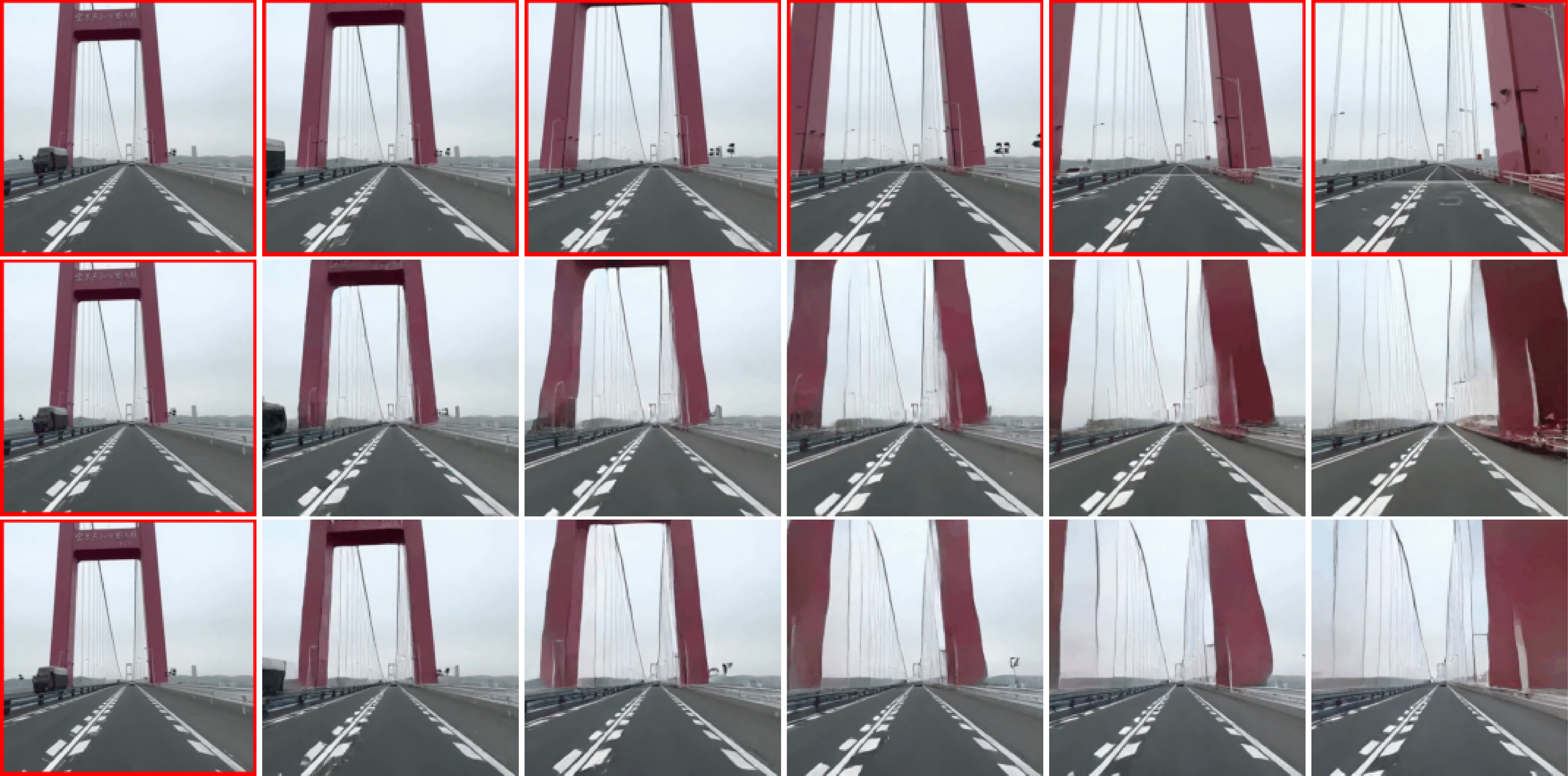}
        \caption{\textit{Lifelong Drive} videos.}
        \label{fig:drive-comparison-road}
        \end{subfigure}
    \end{minipage}
    \hfill
    \begin{minipage}[t]{0.49\linewidth}
        \begin{subfigure}{\linewidth}
        \centering
        \includegraphics[width=6.1cm, height=3.05cm, keepaspectratio=false]{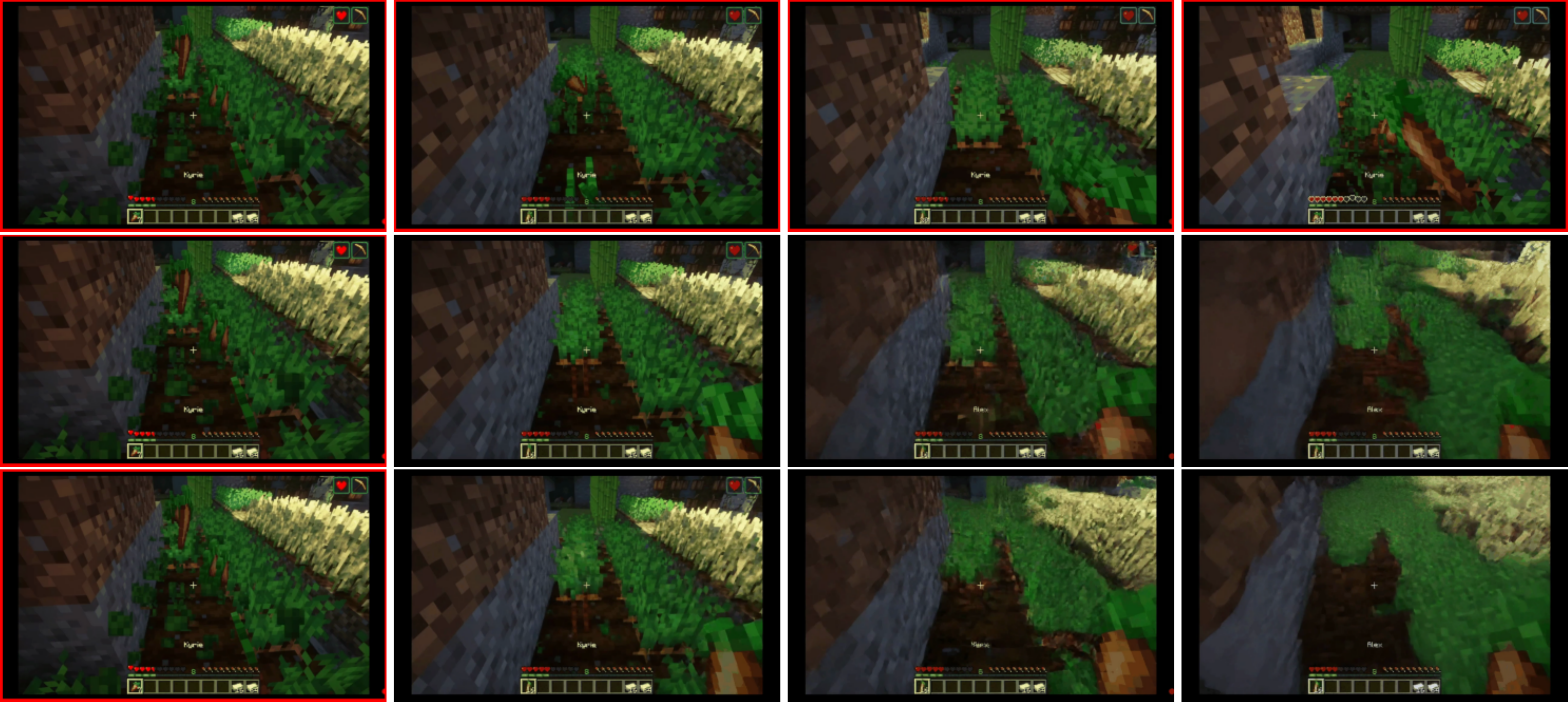}
        \caption{\textit{Lifelong PLAICraft} videos.}
        \label{fig:minecraft-comparison-carrot}
        \end{subfigure}
    \end{minipage}
    \caption{Qualitative result comparisons. The top, middle, and bottom rows depict videos from the ground truth data, offline learned model, and lifelong learned model respectively.}
    \label{fig:all-comparisons-appendix}
\end{figure}

\section{Additional Quantitative Results}
\label{appendix:quantitative}

This section presents the train and test video stream performances of offline training (Offline Learning), lifelong learning using experience replay with limited buffer size (Experience Replay)\footnote{This configuration is referred to as Lifelong Learning in the main text's \Cref{sec:experiments}.}, lifelong learning using experience replay with unlimited buffer size (Full Replay), and lifelong learning without the use of a replay buffer (No Replay) for U-Net models on all datasets.

We make two high-level observations.
Firstly, No Replay cannot effectively learn from datasets that are not largely stationary due to forgetting, even on simple datasets like \textit{Lifelong Bouncing Balls (C)}.
This suggests that when lifelong learning on video streams that contain non-repeating details, mechanisms that preserve past knowledge are necessary, regardless of how simple the video streams might be.
Secondly, we find that Experience Replay and Full Replay performances are not substantially different across the benchmarks, 
suggesting that replay buffers are particularly beneficial for nonstationary datasets and that larger buffer size tends to be better, but increasing the buffer size beyond a certain point does not yield significant gains.

\begin{table*}
    \centering
    \caption{\textit{Lifelong Bouncing Balls (O)} metrics for experience replay with different replay buffer sizes for U-Net models. The metrics are computed from 2 training and 3 sampling seeds.}
    \label{tab:ball_stn_full}
    \ra{1.1}
    \small
    \setlength{\tabcolsep}{3pt}
    \begin{adjustbox}{width=\textwidth}
    \begin{tabular}{@{}*{9}{c}@{}}
        \toprule
        & \multicolumn{4}{c}{\textit{Train Stream}} & \multicolumn{4}{c}{\textit{Test Stream}}\\
        \cmidrule(lr){2-5} \cmidrule(lr){6-9}
        Method & FVD & Loss$^{\times \! 10^{-\!5}}$ & minADE & ColorKL & FVD & Loss$^{\times \! 10^{-\!5}}$ & minADE & ColorKL \\
        \midrule

Offline Learning & 4.5 \tiny{±0.1} & 6.3 \tiny{±0.1} & 1.74 \tiny{±0.03} & 0.006 \tiny{±0.001} & 4.7 \tiny{±0.2} & 6.2 \tiny{±0.2} & 1.71 \tiny{±0.06} & 0.006 \tiny{±0.001} \\

\midrule

No Replay & 4.7 \tiny{±0.1} & 6.7 \tiny{±0.1} & 1.91 \tiny{±0.01} & 0.006 \tiny{±0.0} & 5.0 \tiny{±0.2} & 6.7 \tiny{±0.1} & 1.84 \tiny{±0.03} & 0.005 \tiny{±0.0} \\

Experience Replay & 4.9 \tiny{±0.2} & 6.3 \tiny{±0.2} & 1.82 \tiny{±0.03} & 0.005 \tiny{±0.000} & 4.7 \tiny{±0.2} & 6.2 \tiny{±0.3} & 1.81 \tiny{±0.03} & 0.005 \tiny{±0.000} \\

Full Replay & 4.7 \tiny{±0.1} & 6.3 \tiny{±0.1} & 1.80 \tiny{±0.01} & 0.004 \tiny{±0.001} & 4.6 \tiny{±0.1} & 6.2 \tiny{±0.0} & 1.76 \tiny{±0.02} & 0.003 \tiny{±0.001} \\
        \bottomrule
    \end{tabular}
    \end{adjustbox}
\end{table*}

\begin{table*}
    \centering
    \caption{\textit{Lifelong Bouncing Balls (C)} metrics for experience replay with different replay buffer sizes for U-Net models. The metrics are computed from 2 training and 3 sampling seeds.}
    \label{tab:ball_nstn_full}
    \ra{1.1}
    \small
    \setlength{\tabcolsep}{2.5pt}
    \begin{adjustbox}{width=\textwidth}
    \begin{tabular}{@{}*{9}{c}@{}}
        \toprule
        & \multicolumn{4}{c}{\textit{Train Stream}} & \multicolumn{4}{c}{\textit{Test Stream}}\\
        \cmidrule(lr){2-5} \cmidrule(lr){6-9}
        Method & FVD & Loss$^{\times \! 10^{-\!5}}$ & minADE & ColorKL & FVD & Loss$^{\times \! 10^{-\!5}}$ & minADE & ColorKL \\
        \midrule

Offline Learning & 5.8 \tiny{±0.3} & 6.5 \tiny{±0.1} & 2.04 \tiny{±0.09} & 0.007 \tiny{±0.002} & 5.9 \tiny{±0.2} & 6.5 \tiny{±0.1} & 2.14 \tiny{±0.10} & 0.007 \tiny{±0.001} \\

\midrule

No Replay & 357.4 \tiny{±1.8} & 2240 \tiny{±110} & 2.61 \tiny{±0.08} & 0.021 \tiny{±0.002} & 343.6 \tiny{±1.2} & 2252 \tiny{±108} & 2.73 \tiny{±0.11} & 0.022 \tiny{±0.0} \\

Experience Replay & 5.0 \tiny{±0.1} & 7.4 \tiny{±0.1} & 2.03 \tiny{±0.00} & 0.005 \tiny{±0.001} & 5.7 \tiny{±0.2} & 7.5 \tiny{±0.1} & 2.06 \tiny{±0.00} & 0.005 \tiny{±0.000} \\

Full Replay & 5.0 \tiny{±0.1} & 8.0 \tiny{±0.1} & 2.08 \tiny{±0.03} & 0.005 \tiny{±0.001} & 5.6 \tiny{±0.2} & 7.9 \tiny{±0.2} & 2.12 \tiny{±0.0} & 0.006 \tiny{±0.001} \\
        \bottomrule
    \end{tabular}
    \end{adjustbox}
\end{table*}

\begin{table*}
    \centering
    \caption{\textit{Lifelong 3D Maze} metrics for experience replay with different replay buffer sizes for U-Net models. The metrics are computed from 1 training and 3 sampling seeds.}
    \label{tab:maze_full}
    \setlength{\tabcolsep}{4.5pt}
    \ra{1.1}
    \centering
    \begin{adjustbox}{width=\textwidth}
    \begin{tabular}{@{}*{7}{c}@{}}
        \toprule
        & \multicolumn{3}{c}{\textit{Train Stream}} & \multicolumn{3}{c}{\textit{Test Stream}}\\
        \cmidrule(lr){2-4} \cmidrule(lr){5-7}
        Method & FVD  & JEDi  & Loss 
        & FVD  & JEDi  & Loss  \\
        \midrule

Offline Learning & 34.2 \tiny{±1.3} & 0.083 \tiny{±0.002} & 0.005 \tiny{±0.0} & 28.5 \tiny{±1.2} & 0.060 \tiny{±0.001} & 0.005 \tiny{±0.0} \\
        \midrule

No Replay & 130.6 \tiny{±0.6} & 0.142 \tiny{±0.002} & 0.009 \tiny{±0.0} & 36.9 \tiny{±1.9} & 0.104 \tiny{±0.004} & 0.006 \tiny{±0.0} \\

Experience Replay & 41.2 \tiny{±0.4} & 0.073 \tiny{±0.006} & 0.006 \tiny{±0.0} & 30.7 \tiny{±0.1} & 0.061 \tiny{±0.003} & 0.006 \tiny{±0.0} \\

Full Replay & 34.0 \tiny{±0.3} & 0.080 \tiny{±0.002} & 0.006 \tiny{±0.0} & 35.2 \tiny{±0.5} & 0.075 \tiny{±0.006} & 0.006 \tiny{±0.0} \\

        \bottomrule
    \end{tabular}
    \end{adjustbox}
\end{table*}

\begin{table*}
    \centering
    \caption{\textit{Lifelong Drive} metrics for experience replay with different replay buffer sizes for U-Net models. The metrics are computed from 1 training and 3 sampling seeds.}
    \label{tab:drive_full}
    \setlength{\tabcolsep}{4.5pt}
    \ra{1.1}
    \centering
    \begin{adjustbox}{width=\textwidth}
    \begin{tabular}{@{}*{7}{c}@{}}
        \toprule
        & \multicolumn{3}{c}{\textit{Train Stream}} & \multicolumn{3}{c}{\textit{Test Stream}}\\
        \cmidrule(lr){2-4} \cmidrule(lr){5-7}
        Method & FVD  & JEDi  & Loss 
        & FVD  & JEDi  & Loss  \\
        \midrule

Offline Learning & 15.3 \tiny{±0.2} & 0.071 \tiny{±0.000} & 0.029 \tiny{±0.0} & 17.9 \tiny{±0.3} & 0.102 \tiny{±0.001} & 0.026 \tiny{±0.0} \\
        \midrule

No Replay & 322.8 \tiny{±1.2} & 0.608 \tiny{±0.002} & 0.041 \tiny{±0.0} & 183.4 \tiny{±1.7} & 0.343 \tiny{±0.000} & 0.029 \tiny{±0.0} \\

Experience Replay (42min) & 29.3 \tiny{±0.1} & 0.131 \tiny{±0.000} & 0.039 \tiny{±0.0} & 33.2 \tiny{±0.4} & 0.166 \tiny{±0.000} & 0.028 \tiny{±0.0} \\

Experience Replay & 17.0 \tiny{±0.2} & 0.072 \tiny{±0.000} & 0.030 \tiny{±0.0} & 21.8 \tiny{±0.3} & 0.106 \tiny{±0.000} & 0.027 \tiny{±0.0} \\

Full Replay & 15.8 \tiny{±0.0} & 0.072 \tiny{±0.000} & 0.030 \tiny{±0.0} & 19.3 \tiny{±0.2} & 0.096 \tiny{±0.000} & 0.026 \tiny{±0.0} \\

        \bottomrule
    \end{tabular}
    \end{adjustbox}
\end{table*}

\begin{table*}
    \centering
    \caption{\textit{Lifelong PLAICraft} metrics for experience replay with different replay buffer sizes for U-Net models. The metrics are computed from 1 training and 3 sampling seeds.}
    \label{tab:minecraft_full}
    \setlength{\tabcolsep}{4.5pt}
    \ra{1.1}
    \centering
    \begin{adjustbox}{width=\textwidth}
    \begin{tabular}{@{}*{7}{c}@{}}
        \toprule
        & \multicolumn{3}{c}{\textit{Train Stream}} & \multicolumn{3}{c}{\textit{Test Stream}}\\
        \cmidrule(lr){2-4} \cmidrule(lr){5-7}
        Method & FVD  & JEDi  & Loss 
        & FVD  & JEDi  & Loss  \\
        \midrule

Offline Learning & 59.7 \tiny{±0.2} & 0.874 \tiny{±0.007} & 0.034 \tiny{±0.0} & 119.8 \tiny{±1.3} & 1.119 \tiny{±0.004} & 0.042 \tiny{±0.0} \\
        \midrule

No Replay & 239.5 \tiny{±1.5} & 2.038 \tiny{±0.014} & 0.047 \tiny{±0.0} & 270.6 \tiny{±1.9} & 2.281 \tiny{±0.004} & 0.050 \tiny{±0.0} \\

Experience Replay (1.4h) & 69.8 \tiny{±0.7} & 0.842 \tiny{±0.004} & 0.036 \tiny{±0.0} & 142.4 \tiny{±0.4} & 1.200 \tiny{±0.003} & 0.045 \tiny{±0.0} \\

Experience Replay & 62.9 \tiny{±0.6} & 0.876 \tiny{±0.003} & 0.034 \tiny{±0.0} & 130.8 \tiny{±1.1} & 1.189 \tiny{±0.003} & 0.042 \tiny{±0.0} \\

Full Replay & 67.8 \tiny{±0.7} & 0.867 \tiny{±0.004} & 0.034 \tiny{±0.0} & 131.6 \tiny{±0.7} & 1.170 \tiny{±0.001} & 0.042 \tiny{±0.0} \\

        \bottomrule
    \end{tabular}
    \end{adjustbox}
\end{table*}

\section{Orthogonal Replay}
\label{appendix:orthogonal_replay}

We denote the AdamW- and Orthogonal-AdamW-optimized experience replay objective as Experience Replay and Orthogonal Replay respectively.

\begin{table*}
    \centering
    \caption{\textit{Lifelong Bouncing Balls (O)} metrics computed from 2 training and 3 sampling seeds. The configurations (U) and (T) denote U-Net and Transformer architectures.}
    \label{tab:ball_stn_orthogonal}
    \ra{1.1}
    \small
    \setlength{\tabcolsep}{2.75pt}
    \begin{adjustbox}{width=\textwidth}
    \begin{tabular}{@{}*{9}{c}@{}}
        \toprule
        & \multicolumn{4}{c}{\textit{Train Stream}} & \multicolumn{4}{c}{\textit{Test Stream}}\\
        \cmidrule(lr){2-5} \cmidrule(lr){6-9}

        Method & FVD  & Loss$^{\times \! 10^{-\!5}}$ & minADE  & ColorKL  & FVD  & Loss$^{\times \! 10^{-\!5}}$ & minADE  & ColorKL  \\
        \midrule

Experience Replay (U) & 4.9 \tiny{±0.2} & 6.3 \tiny{±0.2} & 1.82 \tiny{±0.03} & 0.005 \tiny{±0.000} & 4.7 \tiny{±0.2} & 6.2 \tiny{±0.3} & 1.81 \tiny{±0.03} & 0.005 \tiny{±0.000} \\

Orthogonal Replay (U) & 4.8 \tiny{±0.1} & 6.4 \tiny{±0.0} & 1.82 \tiny{±0.02} & 0.006 \tiny{±0.000} & 4.9 \tiny{±0.3} & 6.2 \tiny{±0.0} & 1.75 \tiny{±0.00} & 0.005 \tiny{±0.000} \\

\midrule

Experience Replay (T) & 5.5 \tiny{±0.1} & 6.6 \tiny{±0.2} & 2.22 \tiny{±0.05} & 0.005 \tiny{±0.001} & 5.0 \tiny{±0.3} & 6.5 \tiny{±0.2} & 2.16 \tiny{±0.06} & 0.005 \tiny{±0.001} \\

Orthogonal Replay (T) & 5.5 \tiny{±0.2} & 6.4 \tiny{±0.2} & 2.17 \tiny{±0.02} & 0.006 \tiny{±0.002} & 5.3 \tiny{±0.1} & 6.3 \tiny{±0.2} & 2.11 \tiny{±0.01} & 0.007 \tiny{±0.001} \\

        \bottomrule
    \end{tabular}
    \end{adjustbox}
\end{table*}

\begin{table*}
    \centering
    \caption{\textit{Lifelong Bouncing Balls (C)} metrics computed from 2 training and 3 sampling seeds. The configurations (U) and (T) denote U-Net and Transformer architectures.}
    \label{tab:ball_nstn_orthogonal}
    \ra{1.1}
    \small
    \setlength{\tabcolsep}{2.75pt}
    \begin{adjustbox}{width=\textwidth}
    \begin{tabular}{@{}*{9}{c}@{}}
        \toprule
        & \multicolumn{4}{c}{\textit{Train Stream}} & \multicolumn{4}{c}{\textit{Test Stream}}\\
        \cmidrule(lr){2-5} \cmidrule(lr){6-9}
        Method & FVD  & Loss$^{\times \! 10^{-\!5}}$ & minADE  & ColorKL  & FVD  & Loss$^{\times \! 10^{-\!5}}$ & minADE  & ColorKL  \\
        \midrule

Experience Replay (U) & 5.0 \tiny{±0.1} & 7.4 \tiny{±0.1} & 2.03 \tiny{±0.00} & 0.005 \tiny{±0.001} & 5.7 \tiny{±0.2} & 7.5 \tiny{±0.1} & 2.06 \tiny{±0.00} & 0.005 \tiny{±0.000} \\

Orthogonal Replay (U) & 5.2 \tiny{±0.1} & 8.2 \tiny{±0.1} & 2.10 \tiny{±0.11} & 0.004 \tiny{±0.001} & 6.0 \tiny{±0.1} & 8.2 \tiny{±0.1} & 2.12 \tiny{±0.07} & 0.004 \tiny{±0.000} \\

\midrule

Experience Replay (T) & 5.3 \tiny{±0.1} & 8.1 \tiny{±0.0} & 2.29 \tiny{±0.03} & 0.006 \tiny{±0.001} & 6.4 \tiny{±0.3} & 8.1 \tiny{±0.0} & 2.27 \tiny{±0.03} & 0.006 \tiny{±0.000} \\

Orthogonal Replay (T) & 5.4 \tiny{±0.2} & 7.7 \tiny{±0.1} & 2.23 \tiny{±0.01} & 0.006 \tiny{±0.001} & 6.0 \tiny{±0.2} & 7.8 \tiny{±0.1} & 2.25 \tiny{±0.00} & 0.006 \tiny{±0.001} \\

        \bottomrule
    \end{tabular}
    \end{adjustbox}
\end{table*}

\begin{table*}
    \centering
    \caption{\textit{Lifelong 3D Maze} metrics computed from 1 training and 3 sampling seeds.}
    \label{tab:maze_orthogonal}
    \ra{1.1}
    \small
    \setlength{\tabcolsep}{4.5pt}
    \begin{adjustbox}{width=\textwidth}
    \begin{tabular}{@{}*{7}{c}@{}}
        \toprule
        & \multicolumn{3}{c}{\textit{Train Stream}} & \multicolumn{3}{c}{\textit{Test Stream}}\\
        \cmidrule(lr){2-4} \cmidrule(lr){5-7}
        Method & FVD  & JEDi  & Loss  & FVD  & JEDi  & Loss \\
        \midrule
    
Experience Replay (U-Net) & 41.2 \tiny{±0.4} & 0.073 \tiny{±0.006} & 0.006 \tiny{±0.0} & 30.7 \tiny{±0.1} & 0.061 \tiny{±0.003} & 0.006 \tiny{±0.0} \\

Orthogonal Replay (U-Net) & 41.3 \tiny{±0.4} & 0.080 \tiny{±0.003} & 0.006 \tiny{±0.0} & 31.8 \tiny{±0.7} & 0.084 \tiny{±0.001} & 0.006 \tiny{±0.0}\\

        \midrule
Experience Replay (Transformer) & 32.1 \tiny{±0.9} & 0.082 \tiny{±0.005} & 0.006 \tiny{±0.0} & 30.2 \tiny{±0.8} & 0.070 \tiny{±0.001} & 0.006 \tiny{±0.0} \\

Orthogonal Replay (Transformer) & 34.5 \tiny{±0.6} & 0.067 \tiny{±0.003} & 0.006 \tiny{±0.0} & 29.7 \tiny{±1.2} & 0.071 \tiny{±0.002} & 0.006 \tiny{±0.0}\\

        \bottomrule
    \end{tabular}
    \end{adjustbox}
\end{table*}

\begin{table*}
    \centering
    \caption{\textit{Lifelong Drive} metrics computed from 1 training and 3 sampling seeds.}
    \label{tab:drive_orthogonal}
    \setlength{\tabcolsep}{4.5pt}
    \ra{1.1}
    \small
    \centering
    \begin{adjustbox}{width=\textwidth}
    \begin{tabular}{@{}*{7}{c}@{}}
        \toprule
        & \multicolumn{3}{c}{\textit{Train Stream}} & \multicolumn{3}{c}{\textit{Test Stream}}\\
        \cmidrule(lr){2-4} \cmidrule(lr){5-7}
        Method & FVD  & JEDi  & Loss 
        & FVD  & JEDi  & Loss  \\
        \midrule

Experience Replay (U-Net) & 17.0 \tiny{±0.2} & 0.072 \tiny{±0.000} & 0.030 \tiny{±0.0} & 21.8 \tiny{±0.3} & 0.106 \tiny{±0.000} & 0.027 \tiny{±0.0} \\

Orthogonal Replay (U-Net) & 18.8 \tiny{±0.1} & 0.078 \tiny{±0.000} & 0.030 \tiny{±0.0} & 25.6 \tiny{±0.2} & 0.125 \tiny{±0.000} & 0.026 \tiny{±0.0}\\

        \midrule

Experience Replay (Transformer) & 10.7 \tiny{±0.1} & 0.023 \tiny{±0.000} & 0.029 \tiny{±0.0} & 12.9 \tiny{±0.1} & 0.039 \tiny{±0.000} & 0.025 \tiny{±0.0} \\

Orthogonal Replay (Transformer) & 11.1 \tiny{±0.1} & 0.027 \tiny{±0.000} & 0.029 \tiny{±0.0} & 13.6 \tiny{±0.2} & 0.044 \tiny{±0.000} & 0.025 \tiny{±0.0}\\
        \bottomrule
    \end{tabular}
    \end{adjustbox}
\end{table*}

\begin{table*}
    \centering
    \caption{\textit{Lifelong PLAICraft} metrics computed from 1 training and 3 sampling seeds.}
    \label{tab:minecraft_orthogonal}
    \ra{1.1}
    \small
    \setlength{\tabcolsep}{4.5pt}
    \begin{adjustbox}{width=\textwidth}
    \begin{tabular}{@{}*{7}{c}@{}}
        \toprule
        & \multicolumn{3}{c}{\textit{Train Stream}} & \multicolumn{3}{c}{\textit{Test Stream}}\\
        \cmidrule(lr){2-4} \cmidrule(lr){5-7}
        Method & FVD  & JEDi  & Loss 
        & FVD  & JEDi  & Loss  \\
        \midrule

Experience Replay (U-Net) & 62.9 \tiny{±0.6} & 0.876 \tiny{±0.003} & 0.034 \tiny{±0.0} & 130.8 \tiny{±1.1} & 1.189 \tiny{±0.003} & 0.042 \tiny{±0.0} \\

Orthogonal Replay (U-Net) & 69.2 \tiny{±1.1} & 0.842 \tiny{±0.006} & 0.035 \tiny{±0.0} & 136.3 \tiny{±0.9} & 1.206 \tiny{±0.001} & 0.043 \tiny{±0.0}\\

        \midrule

Experience Replay (Transformer) & 21.9 \tiny{±0.4} & 0.197 \tiny{±0.003} & 0.025 \tiny{±0.0} & 26.3 \tiny{±0.3} & 0.208 \tiny{±0.001} & 0.024 \tiny{±0.0} \\

Orthogonal Replay (Transformer) & 21.9 \tiny{±0.1} & 0.182 \tiny{±0.002} & 0.024 \tiny{±0.0} & 26.0 \tiny{±0.3} & 0.192 \tiny{±0.001} & 0.024 \tiny{±0.0}\\

        \bottomrule
    \end{tabular}
    \end{adjustbox}
\end{table*}

\clearpage

\section{Optimization Time Performance Metrics}
\label{appendix:online_test}

\begin{center}
\begin{minipage}{\linewidth}
    \captionsetup{type=figure}
    \centering
    \begin{subfigure}[c]{\linewidth}
        \centering
        \includegraphics[width=0.9\linewidth]{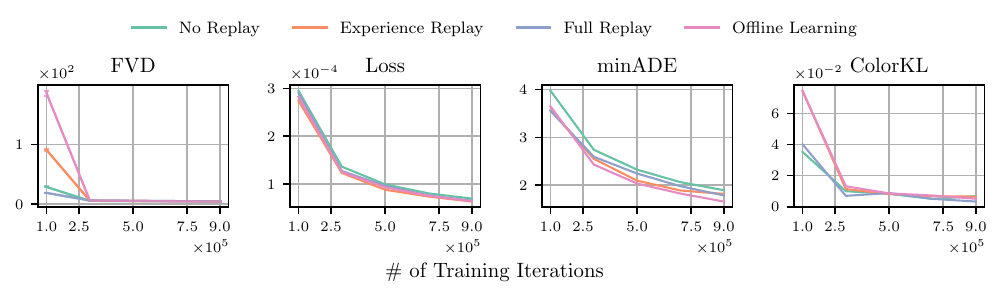}
        \caption{\textit{Lifelong Bouncing Balls (O)} dataset.}
        \label{fig:online_test_ball_stn_appendix}
    \end{subfigure}
    \begin{subfigure}[c]{\linewidth}
        \centering
        \includegraphics[width=0.9\linewidth]{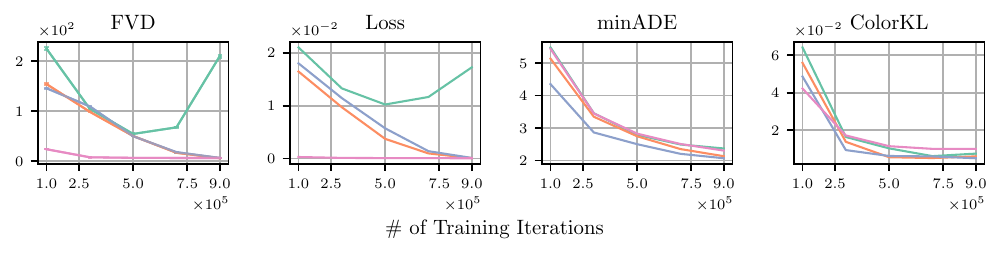}
        \caption{\textit{Lifelong Bouncing Balls (C)} dataset.}
        \label{fig:online_test_ball_nstn_appendix}
    \end{subfigure}
    \begin{subfigure}[c]{\linewidth}
        \centering
        \includegraphics[width=0.9\linewidth]{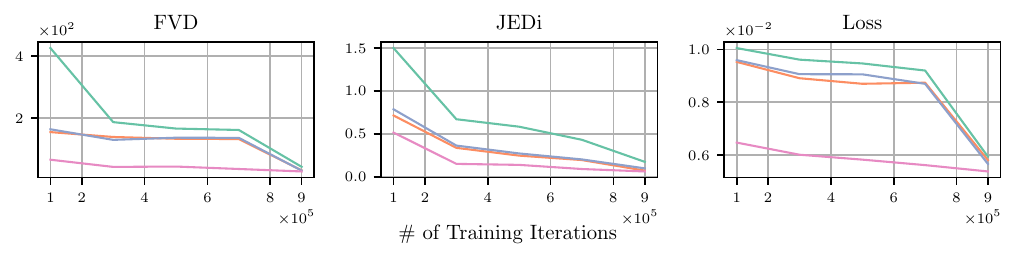}
        \caption{\textit{Lifelong 3D Maze} dataset.}
        \label{fig:online_test_maze_appendix}
    \end{subfigure}
    \begin{subfigure}[c]{\linewidth}
        \centering
        \includegraphics[width=0.9\linewidth]{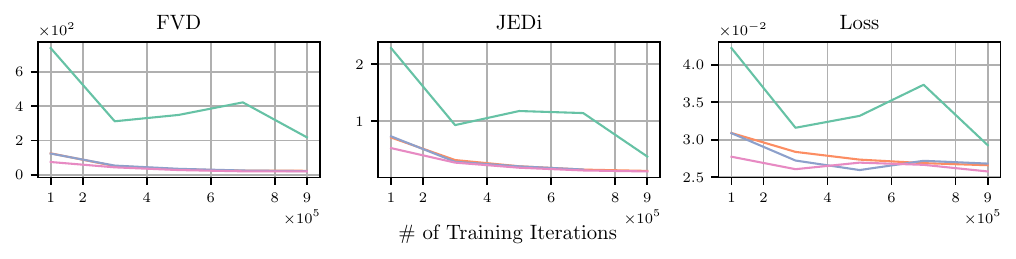}
        \caption{\textit{Lifelong Drive} dataset.}
        \label{fig:online_test_drive_appendix}
    \end{subfigure}
    \begin{subfigure}[c]{\linewidth}
        \centering
        \includegraphics[width=0.9\linewidth]{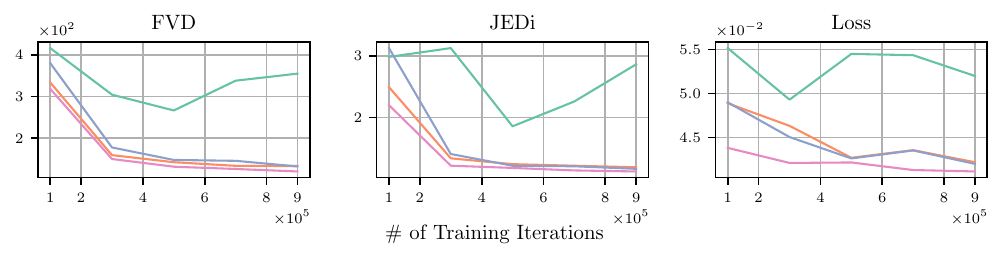}
        \caption{\textit{Lifelong PLAICraft} dataset.}
        \label{fig:online_test_minecraft_appendix}
    \end{subfigure}
    \addtocounter{figure}{-1}
    \captionof{figure}{
    Test stream performance metrics for the U-Net model checkpoints at different training iterations (refer to \Cref{appendix:quantitative} for baseline details).
    The plots show the improvement in model quality as online training progresses.
    All models generally improve the longer they are trained.}
    \label{fig:online_test_plots_appendix}
\end{minipage}
\end{center}

\clearpage

\section{Train Stream Performance Breakdown}
\label{appendix:timeline}

\begin{center}
\begin{minipage}{\linewidth}
    \captionsetup{type=figure}
    \centering
    \begin{subfigure}[c]{\linewidth}
        \centering
        \includegraphics[width=0.9\linewidth]{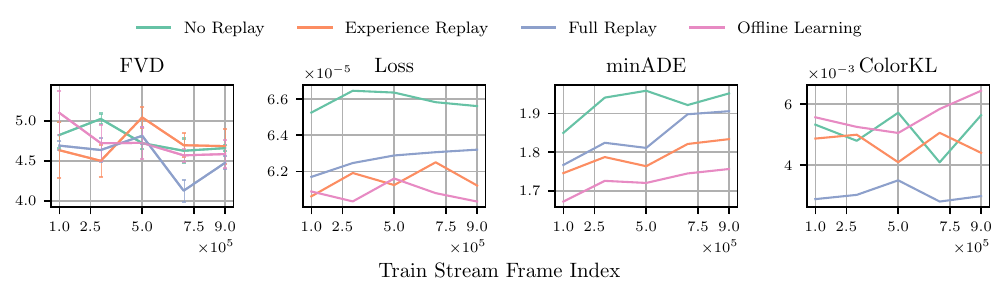}
        \caption{\textit{Lifelong Bouncing Balls (O)} dataset.}
        \label{fig:timeline_ball_stn}
    \end{subfigure}
    \begin{subfigure}[c]{\linewidth}
        \centering
        \includegraphics[width=0.9\linewidth]{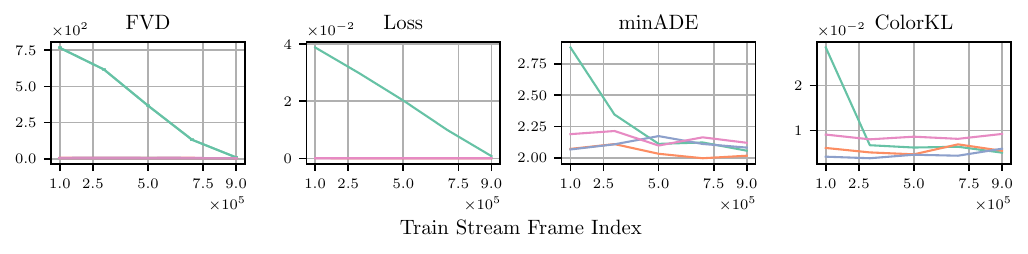}
        \caption{\textit{Lifelong Bouncing Balls (C)} dataset.}
        \label{fig:timeline_ball_nstn}
    \end{subfigure}
    \begin{subfigure}[c]{\linewidth}
        \centering
        \includegraphics[width=0.9\linewidth]{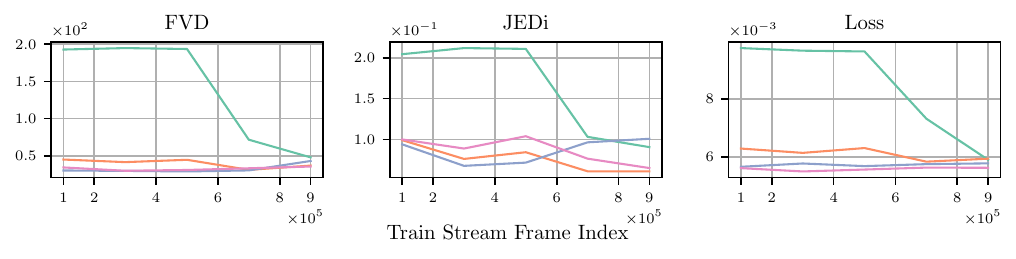}
        \caption{\textit{Lifelong 3D Maze} dataset.}
        \label{fig:timeline_maze}
    \end{subfigure}
    \begin{subfigure}[c]{\linewidth}
        \centering
        \includegraphics[width=0.9\linewidth]{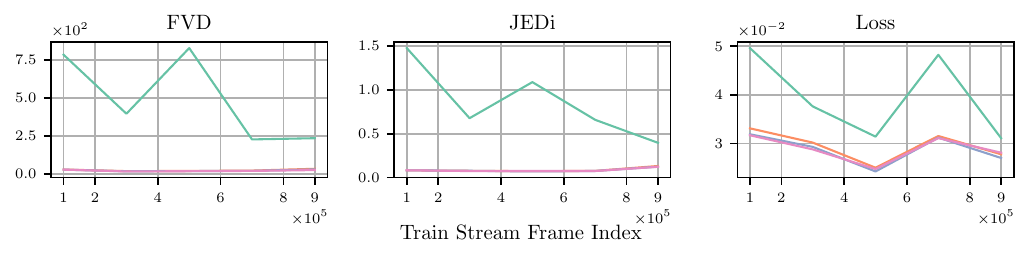}
        \caption{\textit{Lifelong Drive} dataset.}
        \label{fig:timeline_drive}
    \end{subfigure}
    \begin{subfigure}[c]{\linewidth}
        \centering
        \includegraphics[width=0.9\linewidth]{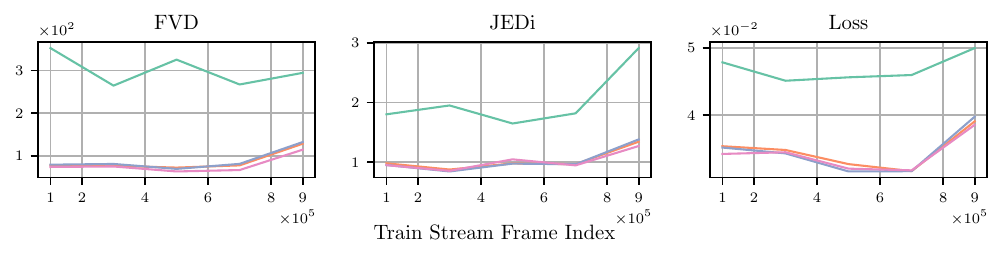}
        \caption{\textit{Lifelong PLAICraft} dataset.}
        \label{fig:timeline_minecraft}
    \end{subfigure}
    \addtocounter{figure}{-1}
    \captionof{figure}{
    Frame generation performance metrics of fully trained U-Net video diffusion models, computed from frames selected from different points in the training stream.
    Each point represents an evaluation on 1,000 consecutive frame subsequences that follow the corresponding training stream frame index.}
    \label{fig:timeline_plots}
\end{minipage}
\end{center}

\section{Human Evaluation of Lifelong Learning vs.\ Offline Learning}
\label{appendix:human_eval}
 
To assess whether lifelong learning degrades perceptual quality relative to standard offline learning, we conducted a human evaluation study.
We aggregated 76 offline learning and lifelong learning clip pairs (4 generations $\times$ 19 scenarios) that illustrate key characteristics of the datasets (e.g., screen flip rare event for Lifelong 3D Maze, lane change for Lifelong Drive, trading UI for Lifelong PLAICraft).
For every pair, five independent Amazon Mechanical Turk participants were asked—without knowing which model produced which clip—to indicate whether the left or right clip was of higher quality, or whether the two appeared similar.
The aggregated preferences across datasets are reported in \cref{tab:human_eval}.
 
\begin{table}
  \centering
  \caption{Human preference results comparing future frames generated by offline and lifelong learned models.
  Five annotators judged 76 clip pairs in total.
  Votes are aggregated across annotators.}
  \label{tab:human_eval}
  \begin{tabular}{lccc}
    \toprule
    Dataset & Offline Better & Lifelong Better & Similar \\
    \midrule
    Lifelong Bouncing Balls (O) & 24 & 25 & 11 \\
    Lifelong Bouncing Balls (C) & 23 & 26 & 11 \\
    Lifelong 3D Maze            & 19 & 25 & 16 \\
    Lifelong Drive              & 43 & 41 & 16 \\
    Lifelong PLAICraft          & 43 & 40 & 17 \\
    \bottomrule
  \end{tabular}
\end{table}
 
Across all datasets, participants show no consistent preference for either offline or lifelong learning. Even when excluding responses marked ``similar'', votes remain balanced between the two methods, indicating that the perceptual quality of lifelong-trained models is on par with their offline-trained counterparts.

\begin{figure}
    \centering
    \includegraphics[width=0.85\linewidth]{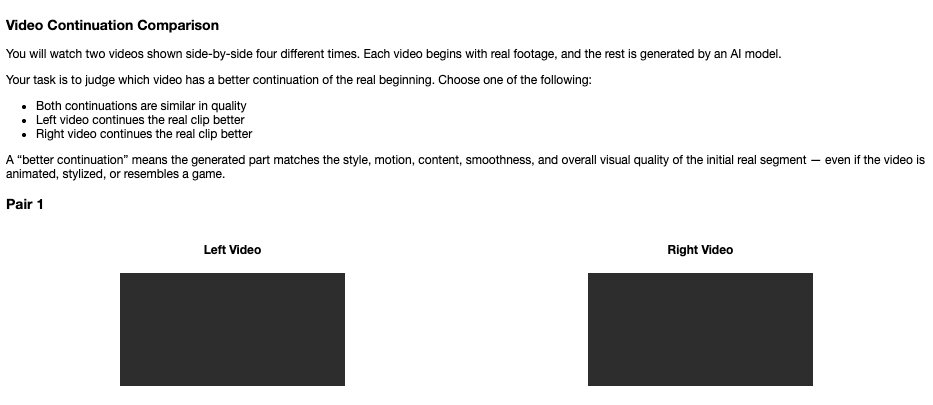}
    \caption{Sample screen shown to the Mechanical Turk participants.}
    \label{fig:placeholder}
\end{figure}

\section{EgoLife Results}
\label{appendix:egolife}

We use the first one million frames of the \emph{A1\_JAKE} participant's EgoLife stream as the training stream and the subsequent 100,000 frames as a held-out test stream.
We compute the test metrics on 1,000 video subsequences selected at evenly spaced positions throughout the test stream.
All other experimental settings match those used for \emph{Lifelong PLAICraft}, including the model architectures and sizes, context length, batch size, and replay-buffer configuration (see \Cref{appendix:expdetails}).

\begin{table*}
    \centering
    \caption{\textit{EgoLife} metrics ($\downarrow$) computed from 1 training and 3 sampling seeds.}
    \label{tab:egolife}
    \setlength{\tabcolsep}{4.5pt}
    \ra{1.05}
    \small
    \centering
    \begin{adjustbox}{width=\textwidth}
    \begin{tabular}{@{}*{7}{c}@{}}
        \toprule
        & \multicolumn{3}{c}{\textit{Train Stream}} & \multicolumn{3}{c}{\textit{Test Stream}}\\
        \cmidrule(lr){2-4} \cmidrule(lr){5-7}
        Method & FVD  & JEDi  & Loss 
        & FVD  & JEDi  & Loss  \\
        \midrule

Offline Learning (U-Net) & 11.8 \tiny{±0.1} & 0.132 \tiny{±0.001} & 0.029 \tiny{±0.0} & 13.9 \tiny{±0.1} & 0.168 \tiny{±0.000} & 0.030 \tiny{±0.0} \\

Lifelong Learning (U-Net) & 12.0 \tiny{±0.0} & 0.142 \tiny{±0.001} & 0.029 \tiny{±0.0} & 14.6 \tiny{±0.1} & 0.193 \tiny{±0.000} & 0.031 \tiny{±0.0} \\

        \midrule

Offline Learning (Transformer) & 11.8 \tiny{±0.0} & 0.052 \tiny{±0.000} & 0.025 \tiny{±0.0} & 12.6 \tiny{±0.0} & 0.078 \tiny{±0.002} & 0.026 \tiny{±0.0} \\

Lifelong Learning (Transformer) & 12.0 \tiny{±0.1} & 0.054 \tiny{±0.000} & 0.026 \tiny{±0.0} & 13.3 \tiny{±0.2} & 0.078 \tiny{±0.001} & 0.027 \tiny{±0.0} \\

        \bottomrule
    \end{tabular}
    \end{adjustbox}
\end{table*}

\end{document}